\documentclass{article}

\usepackage{microtype}
\usepackage{graphicx}
\usepackage{subfigure}
\usepackage{booktabs} 

\usepackage{amsfonts, amsmath, amsthm, amssymb}

\usepackage[shortlabels]{enumitem}
\setitemize{fullwidth}

\usepackage{afterpage}

\usepackage{hyperref}
\usepackage[warn]{textcomp}

\usepackage[accepted]{icml2020}

\icmltitlerunning{Deep Ensembles on a Fixed Memory Budget: One Wide Network or Several Thinner Ones?
}

\begin{document}

\twocolumn[
\icmltitle{Deep Ensembles on a Fixed Memory Budget:\\ One Wide Network or Several Thinner Ones?
}



\icmlsetsymbol{equal}{*}

\begin{icmlauthorlist}
\icmlauthor{Nadezhda Chirkova}{hse}
\icmlauthor{Ekaterina Lobacheva}{hse}
\icmlauthor{Dmitry Vetrov}{hse,ss}
\end{icmlauthorlist}

\icmlaffiliation{hse}{Samsung-HSE Laboratory, National Research University Higher School of Economics, Moscow, Russia}
\icmlaffiliation{ss}{Samsung AI Center Moscow, Russia}

\icmlcorrespondingauthor{Nadezhda Chirkova}{nchirkova@hse.ru}
\icmlcorrespondingauthor{Ekaterina Lobacheva}{elobacheva@hse.ru}

\icmlkeywords{Machine Learning, deep ensembles, memory budget}

\vskip 0.3in
]



\printAffiliationsAndNotice{} 

\begin{abstract}

One of the generally accepted views of modern deep learning is that increasing the number of parameters usually leads to better quality. The two easiest ways to increase the number of parameters is to increase the size of the network, e.\,g. width, or to train a deep ensemble; both approaches improve the performance in practice. In this work, we consider a fixed memory budget setting, and investigate, what is more effective: to train a single wide network, or to perform a \emph{memory split} --- to train an ensemble of several thinner networks, with the same total number of parameters? We find that, for large enough budgets, the number of networks in the ensemble, corresponding to the optimal memory split, is usually larger than one.  Interestingly, this effect holds for the commonly used sizes of the standard architectures. For example, one WideResNet-28-10 achieves significantly worse test accuracy on CIFAR-100 than an ensemble of sixteen thinner WideResNets: 80.6\% and 82.52\%  correspondingly. We call the described effect the Memory Split Advantage and show that it holds for a variety of datasets and model architectures.

\end{abstract}

\section{Introduction}
\begin{figure}[h!]
\vskip 0.2in
\begin{center}
\centerline{
\includegraphics[height=4.5cm]{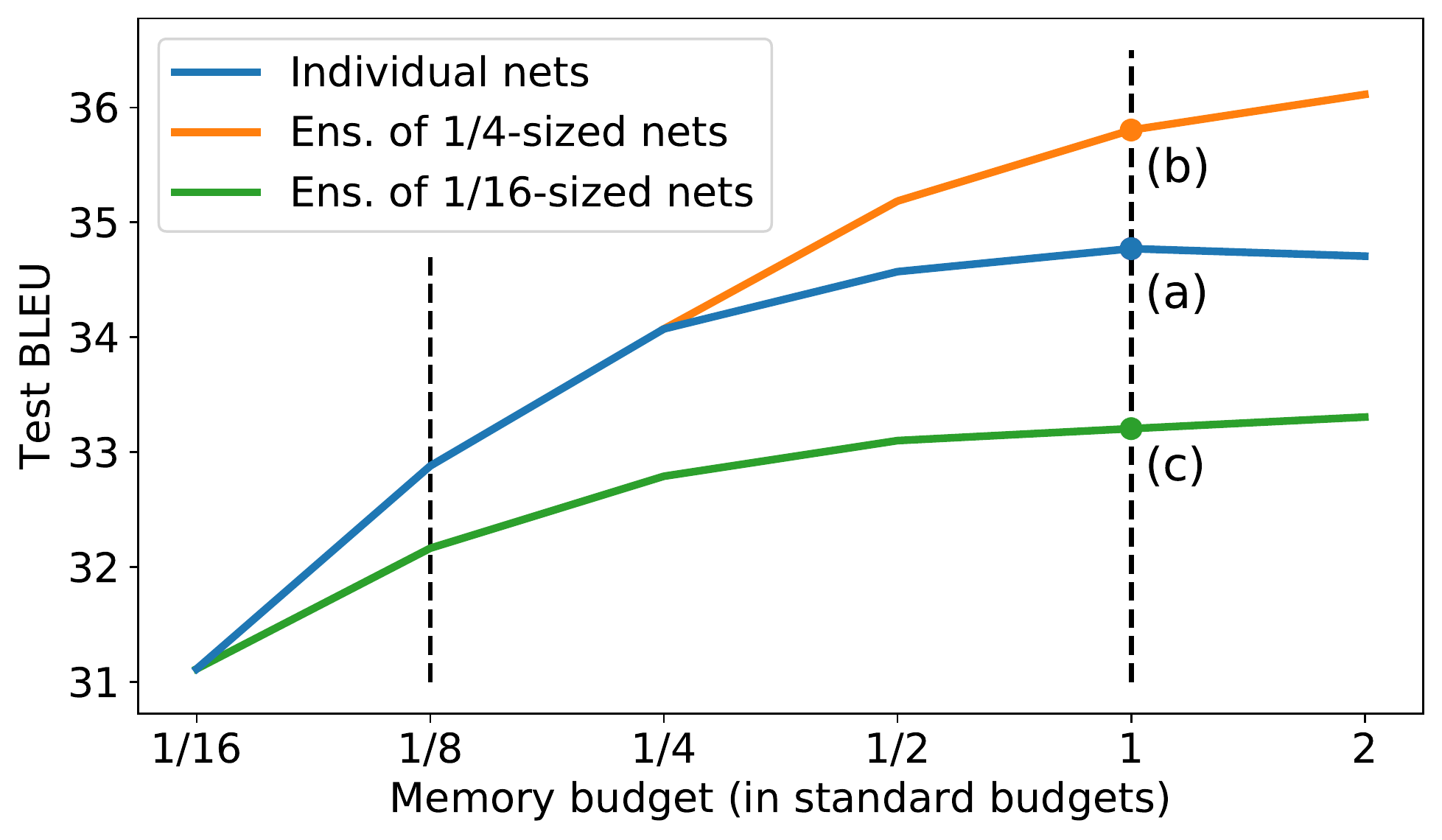}
}
\caption{Illustration of the MSA effect. The relative arrangement of the quality curves of an individual model and ensembles leads to the MSA effect for large enough budgets: the ensemble of several medium-width neural networks (b) gives higher quality than one very wide neural network (a) or the ensemble of a lot of thin neural networks (c). The results for Transformer on IWSLT'14 De-En are shown, x-axis denotes the total number of parameters, standard budget correspond to the size of Transformer with $d_{model}=512$.}
\label{fig:motivation}
\end{center}
\vskip -0.2in
\end{figure}
In recent decade, a lot of neural network architectures for solving different artificial intelligence problems were proposed, and each new state-of-the-art model usually has more parameters than its predecessor.
While working within a fixed architectural family, such as WideResNets~\citep{wrn} or Transformers~\citep{transformer}, a variety of upscaling methods can be used to increase the network size. The two most common approaches are to increase the model depth, i.\,e. the number of layers~\citep{deep_resnet,deep_transformer,deep_speech}, or to increase the model width by, for example, increasing the number of filters in a convolutional neural network or the dimensionality of embeddings and fully-connected layers in a Transformer. 
Both methods improve model performance and also can be combined together for higher effectiveness~\citep{efficient_net}. In this work, we focus on increasing the network size by scaling the width of each layer because straightforward upscaling of the network depth often leads to optimization difficulties~\citep{stoch_depth,deep_transformer_training}.

Rather than increasing the size of one model, one can train an ensemble of several models. One of the most popular methods to construct an ensemble of deep neural networks is to train individual networks from different random initializations, and then average their predictions~\citep{deepens90,deepens}. Such ensembles are called Deep Ensembles. Deep Ensembles with an increasing ensemble size were shown to improve both classification performance and quality of uncertainty estimation of the model~\citep{inception,google_translate,bert,pitfalls}.

The effect of increasing the network size or the ensemble size independently is well investigated in the literature. In this work, we investigate this effect in a fixed memory budget setting, when increasing the network size entails decreasing  the ensemble size. Under a fixed memory budget, we mean a fixed number of parameters.
We focus on the following question: with a fixed number of parameters, what performs better: (a) one very wide neural network, or (b) the ensemble of several medium-width neural networks, or (c) the ensemble of a lot of thin neural networks? 
Our main empirical result is that, for large enough memory budgets, (b) is better than (a) and (c): splitting the memory budget between several mid-size neural networks results in better performance than spending the budget on one big network or training a huge number of small networks (see figure~\ref{fig:motivation}). We call this effect Memory Split Advantage (MSA).
We perform a rigorous empirical study of the MSA effect and show the effect for VGG~\citep{vgg} and WideResNet on CIFAR-10/100 datasets as well as Transformer on IWSLT‘14 German-to-English.
We observe that for a lot of dataset--archirecture pairs, the MSA effect holds even for small architecture configurations, i.\,e.\, several times smaller than commonly used ones.

The MSA-effect leads to a simple and effective way of improving the quality of the model without changing the number of parameters. Splitting the network into an ensemble of several smaller ones allows distributed training and prediction.

\section{Related work}
\paragraph{Individual networks.}
According to the conventional bias-variance trade-off theory~\citep{hastie}, for a particular task and model family, there should be an optimal model size, so that larger, more complex, models, overfit and perform worse. However, a tendency of larger neural networks to achieve higher performance is observed in many practical applications of modern deep learning~\citep{gpipe,efficient_net,gpt2,bert}, even though such models are overparametrized and are able to fit even random labels~\citep{randomlab}. This phenomena is actively researched nowadays and recent works~\citep{bvd, openai} confirm that, in an overparametrized regime, increasing the size of the network leads to a better quality.

\paragraph{Ensembles.}
The standard ensembling approach for neural networks consists in training individual models independently and then averaging their predictions. A diversity in error distributions across member networks is essential to construct an effective ensemble~\citep{deepens90,deepens90_2}. To achieve it, networks are usually trained on the same dataset but from different random initializations~\citep{deepens}.
This approach results in higher performance than standard bagging in case of neural networks~\citep{mheads}. 
Ensembling of neural networks was shown to substantially increase both classification performance and quality of uncertainty estimation in a lot of practical tasks~\citep{inception,google_translate,bert,ensembles_active_learning,pitfalls}. Moreover, a lot of winning and top performing solutions of different Kaggle competitions\footnote{\url{www.kaggle.com}} use ensembles of deep neural networks.

\paragraph{Memory efficient ensembles.}
While achieving higher performance, ensembles of neural networks are also more resource consuming in terms of memory.
Various compression methods may be used to compress individual networks, such as sparsification and quantization~\citep{deep_compression} or more compact parametrization of weight matrices~\citep{tensor_train}. To compress an ensemble, one may compress each member network independently or apply, for example, distillation~\citep{distill} to the whole ensemble. These approaches and our findings are orthogonal and can be straightforwardly combined. 

More compact ensembles may also be constructed by sharing parameters between member networks~\citep{mheads,intra_ens}. \citet{intra_ens} propose a technique of simultaneous training of several sub-networks inside one network so that these sub-networks are then used as an ensemble. Hence, the authors also split one network into an ensemble, but in a different way: the sub-networks are dependent and share a large portion of parameters, while our findings show that even splitting one network into several independent ones boosts the performance for large enough budgets.

Multi-branch architectures~\cite{inception, resnext}, which split some building blocks of a network into sub-blocks and aggregate (e. g. sum or concatenate) sub-blocks' outputs, can also be seen as an ensemble of a large number of subnetworks. These subnetworks are again closely connected and trained jointly, while in our work, we investigate the effect of independent training of networks in the ensemble.

\section{Motivation}
\label{sec:motiv}
In this paper, we are focused on solving a machine learning task by an ensemble, of one or more neural networks, with a fixed total number of parameters. For a fixed memory budget, we consider ensembles containing different number of networks and vary the width
of member networks, so that the resulting combination of the two parameters, number of networks and network's width, results in the right memory budget. 
{ We call the described ensembles, with a fixed number of parameters,  the memory splits.}
We do not change any other architectural properties of individual networks except their width. 

The dependency of individual neural network's performance on its width usually looks as shown in figure~\ref{fig:motivation}: quality increases when the width grows and saturates for larger widths~\citep{efficient_net}. As a result, in practice, it is preferable to use a network, wide enough to achieve high performance, but not too wide, because it needs a lot of resources while providing only a negligible quality improvement. 

The same type of dependency is observed between quality and number of networks in Deep Ensembles: the more networks the better, but for the large number of networks quality saturates~\citep{pitfalls}. 
An ensemble quality curve (the network size is fixed, the ensemble size varies)
may be depicted in the same plot as the quality curve of an individual network, by using the number of parameters on the x-axis (instead of width or number of networks). We show this curve for ensembles with member networks of two sizes in figure~\ref{fig:motivation}.

When working with a fixed memory budget, one needs to choose how to spend it: to train one large network or to split the budget into two or more parts and train several smaller networks. Intuitively, the choice depends on the budget and relative arrangement of the quality curves of an individual model and ensembles. For a small budget (left vertical line in figure~\ref{fig:motivation}), one model is expected to give higher performance, because the curve for one model increases faster than for an ensemble. We assume such behavior on the premise that smaller models have a high bias, while ensembling tends to decrease the variance~\citep{variance}. For a large budget (the right vertical line in figure~\ref{fig:motivation}), the performance of an individual network saturates, while the common practices say that ensembling increases performance even for very large networks~\citep{inception,bert}. 
Based on these observations, we hypothesize that for large budgets the MSA effect holds: it is preferable to split the budget and train an ensemble of several networks, instead of one large network. 

In this paper, we consider several dataset--architecture pairs and empirically investigate the following research questions (RQs):
\begin{enumerate}[labelindent=\parindent,leftmargin=*,label=RQ\arabic*]
    \item Does the MSA effect hold for various datasets and architectures? 
    \item For which budgets does it hold? Does it hold for standard budgets that are usually used in practice?
    \item How does an optimal split look like for different budgets?
\end{enumerate}

\section{Experimental design}
We perform experiments with convolutional neural networks (WideResNet28x10 and VGG16) on CIFAR-10~\cite{CIFAR10} and CIFAR-100~\cite{CIFAR100}, and Transformer on IWSLT’14 German-English (De-En)~\cite{iwslt}. For each dataset--architecture pair, we investigate memory splitting for a range of memory budgets. 
To obtain different memory splits, we select the width factor, a hyperparameter controlling layer widths, so that the size of the member network is approximately equal to the budget divided by the ensemble size.

Since the good performance of the model is usually obtained with a carefully tuned hyperparameters, and the optimal hyperparameters may change for different network sizes, we investigate memory splitting in two settings: 
\begin{enumerate}[(A)]
    \item without hyperparameter tuning: we use the same hyperparameters for all memory splits, including turning off the regularization (weight decay, dropout);
    \item with hyperparameter tuning: we use grid search to tune the hyperparameters of a single network on the validation set for each network size; to train memory splits we use hyperparameters, optimal for member networks.
\end{enumerate}

We can only find good hyperparameters approximately, introducing additional noise to the results. 
Setting A avoids this additional noise and results in more smooth plots. Setting A is also similar to the setting of~\cite{openai}.
Setting B is more practically oriented.

For each dataset--architecture pair, in each setting, 
we
provide a memory split plot.

\paragraph{Memory split plot.}
Each plot contains several lines, one line corresponds to a fixed memory budget. For a budget $B$ (the number of parameters), we train several memory splits, each contains $N$ networks of size $B/N$, $N=1, 2, 4, 8\dots$ (logarithmic scale). To obtain a network of size $B/N$, we adjust the width factor of the network. 
We then plot the test quality vs. ensemble size $N$ and analyze what is the optimal $N$. An ensemble, corresponding to the optimal $N$, is called optimal memory split for budget $B$. Optimal $N$ usually varies for different budgets $B$.

For the majority of points on the plot, we average the test quality over 3--5 runs (3 for more computationally expensive runs, 5 --- for less expensive ones), and plot mean $\pm$ standard deviation of quality. The most computationally expensive ensembles ($N \geqslant 8$ for one or two biggest budgets) were trained only once, but the standard deviation of their test quality is small because of the large ensemble size $N$.

In the rest of the paper, we denote an ensemble of $N$ networks, each having $S$ parameters, with $E(N, S)$. The total number of parameters in $E(N, S)$ equals $N S$. $Q(E(N, S))$ denotes the test quality of ensemble $E(N, S)$. 
$B_{standard}$ denotes the standard budget, equivalent to the number of parameters in a single network of a commonly used size for the specific architecture.

\afterpage{
\begin{table}[t]
\vskip -0.1in
\caption[Hyperparameter choices for all architectures and settings. Notation: wd --- weight decay, lr --- initial learning rate, dr --- dropout rate. In setting A, we use fixed hyperparameters for all width factors. In setting B, for each width factor, we choose optimal hyperparameters on the validation set using grid search (all possible combinations).]{Hyperparameter choices for all architectures and settings. Notation: wd --- weight decay, lr --- initial learning rate, dr --- dropout rate. In setting A, we use the fixed hyperparameters for all network sizes. In setting B, for each network size, we choose the optimal hyperparameters on the validation set using grid search (all possible combinations\footnotemark).}
\label{tab:hyper}
\begin{center}
\begin{small}
\begin{tabular}{lcr}
\toprule
Architecture & A: without  & B: with\\
            & hyp. tuning & hyp. tuning  \\
\midrule
VGG & lr = $0.005$ &  lr $\in \{ 0.005, 0.05 \}$ \\
    & wd = $0$ &  wd $\in \{10^{-4}, 3\cdot10^{-4},$ \\
    && $10^{-3}, 3\cdot 10^{-3}\}$ \\
    & dr = $0$ & dr $\in \{ 0, 0.25, 0.5 \}$ \\
\midrule
WideResNet & lr = $0.01$ &  lr $\in \{ 0.01, 0.1 \}$ \\
           & wd = $0$ &  wd $\in \{10^{-4}, 3\cdot10^{-4},$ \\
    && $10^{-3}, 3\cdot 10^{-3}\}$ \\
\midrule
Transformer & lr = $0.0004$ & lr $\in \{\text{from } 10^{-4} \text{  to  } 5\cdot 10^{-3},$\\ && step $= 10^{-4}  \}$ \\
            & wd = $0$ & wd $\in \{10^{-i} \text{ for } i = \overline{1,5}  \}$ \\
            & dr = $0$ & dr $\in \{\text{from } 0 \text{ to } 0.4,$\\ && step $= 0.1  \}$\\
\bottomrule
\end{tabular}
\end{small}
\end{center}
\vskip -0.1in
\end{table}
\footnotetext{Due to the limited computational resources, for Transformer, we first select the optimal lr and dr from all possible combinations with wd$=10^{-4}$, and then select the optimal wd.}
}

\paragraph{Experimental details for CNNs.}  We consider VGG~\citep{vgg} (16 layers) and WideResNet~\citep{wrn} (28 layers). We use the implementation provided by ~\citet{fge}\footnote{\url{https://github.com/timgaripov/dnn-mode-connectivity}}, and scale the number of filters for convolutional layers and the number of neurons for fully-connected layers. For VGG\,/\,WideResNet, we use convolutional layers with $[k, 2k, 4k, 8k]$\,/\, $[k, 2k, 4k]$ filters, and fully-connected layers with $8k$\,/\,$4k$ neurons, and vary width factor $k$. 
For VGG, we consider $2 \leqslant k \leqslant 181$, $k=64$ corresponds to a standard, commonly used, configuration. We use the size of this configuration (15.3M parameters) as a standard budget $B_{standard}$ in the experiments with VGG. For WideResNet, we consider $5 \leqslant k \leqslant 453$, $k=160$ corresponds to a standard model, WideResNet-28-10. We use the size of this configuration (36.8M parameters) as a standard budget $B_{standard}$ in the experiments with WideResNet.

We train all the networks for 200 epochs with SGD with an annealing learning schedule and a batch size of 128. In setting A, we use an initial learning rate 10 times smaller than in the reference implementation to ensure that the training converges for all considered models. In setting B, we consider both the standard learning rate and the small one.  
For VGG, we use weight decay, and binary dropout for fully-connected layers.
For WideResNet, we use weight decay 
and batch normalization~\cite{batchnorm}. Dropout does not affect quality a lot, so we do not use it for WideResNet, to reduce the hyperparameter grid size. Hyperparameter choices for all cases are listed in table~\ref{tab:hyper}. Quality is measured in accuracy. More details are given in Appendix~\ref{append:details}.

\paragraph{Experimental details for Transformer.} 
We use standard Transformer architecture~\citep{transformer} with 6 layers, each with model dimensions $d_{model} = k$, feed-forward dimensions $d_{ffn} = 2k$, and 4 attention heads. We use $32 \leqslant k \leqslant 1048$ to obtain neural networks with different memory budgets, $k=512$ corresponds to a standard, commonly used, configuration. We use the size of this configuration (39.5M parameters) as a standard budget in the experiments with Transformer. We use the implementation provided by fairseq\footnote{\url{https://github.com/pytorch/fairseq}}~\citep{fairseq}.

We use label smoothing of $0.1$ and batches of maximum 4096 tokens. We train models using Adam~\citep{adam} with $\beta_1 = 0.9, \beta_2 = 0.98$ and inverse square-root learning rate schedule with 4000 warm-up steps. We stop training after the convergence of the validation loss or after 100/50 epochs for setting A/B, whichever comes first.
In setting A, we use a learning rate of $0.0004$ to ensure the stable training of all network sizes. 
In setting B, we use a weight decay and a binary dropout, and choose optimal hyperparameters using grid search (see table~\ref{tab:hyper}). For evaluation, we use a beam search with a beam size of 5 and a length penalty of 1.0. The translation quality is measured in BLEU~\citep{bleu}. More details are given in Appendix~\ref{append:details}.

\section{Memory split advantage effect}
\begin{figure*}[h!]
\vskip 0.1in
\begin{center}
\centerline{
 \begin{tabular}{cc}
        \small{WideResNet, CIFAR-100, no hyperparameter tuning} & \small{WideResNet, CIFAR-100, with tuned hyperparameters} \\
        \includegraphics[height=4.5cm]{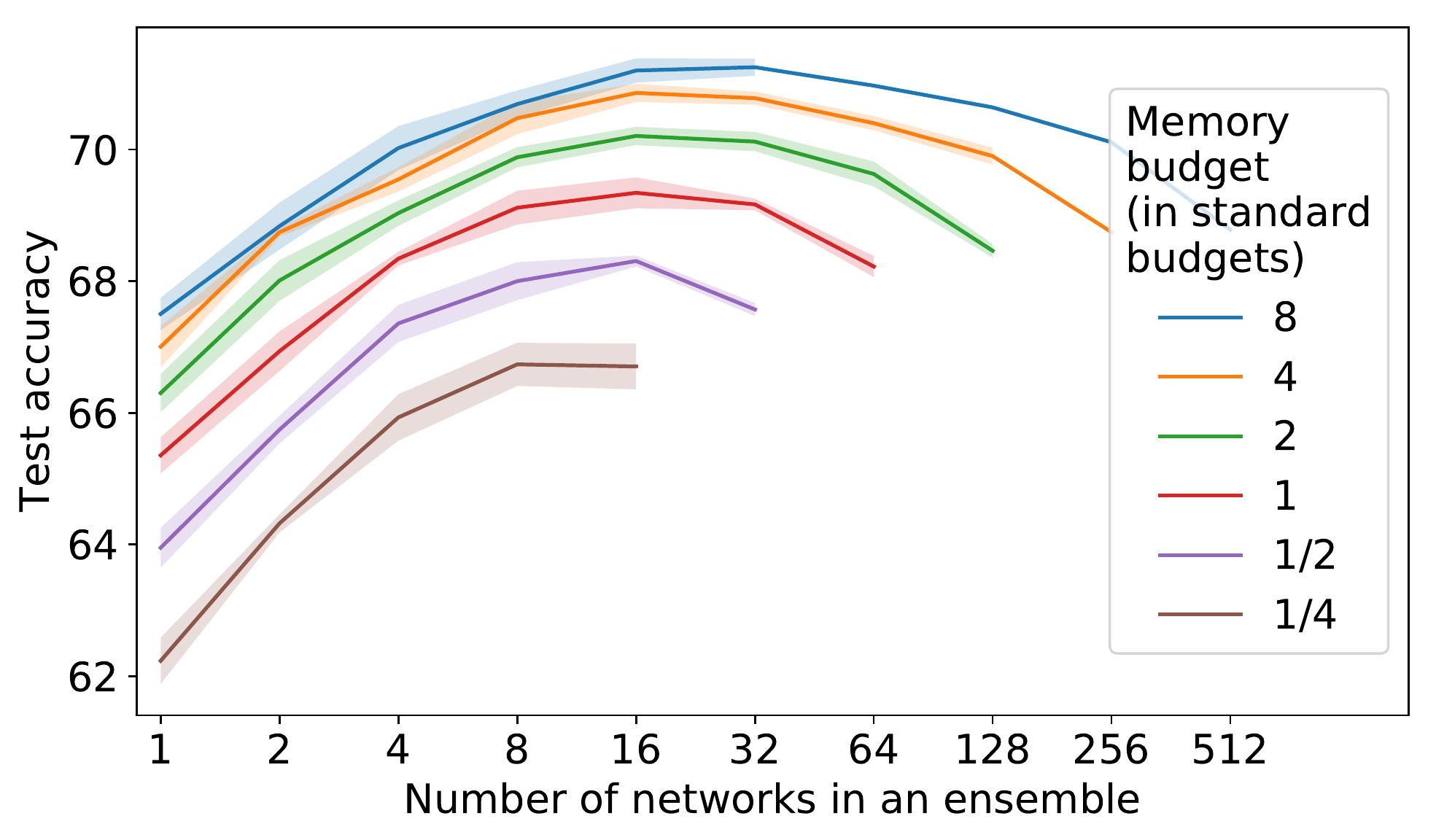}
           & \includegraphics[height=4.5cm]{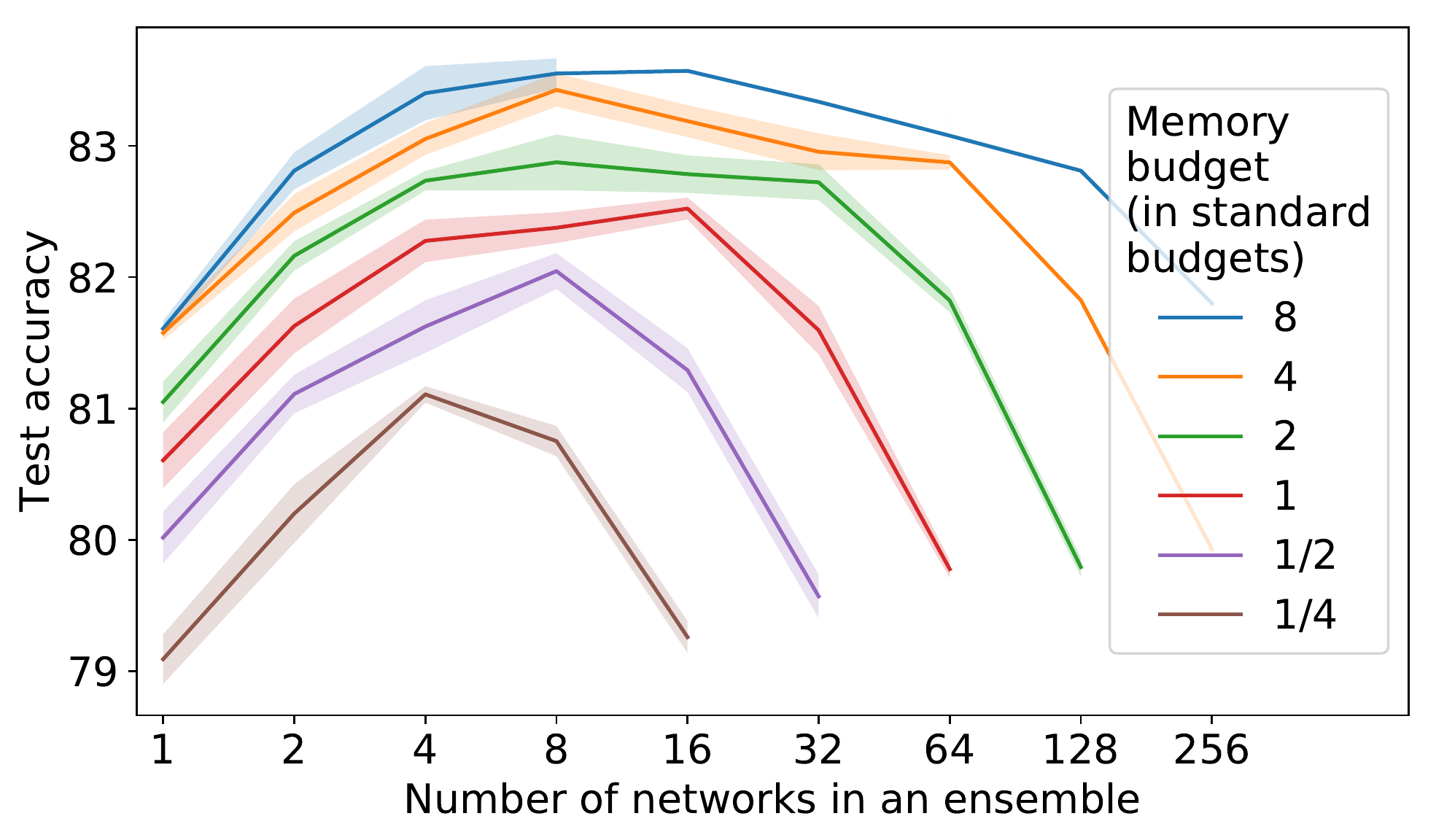} \\
        \small{VGG, CIFAR-100, no hyperparameter tuning} & \small{VGG, CIFAR-100, with tuned hyperparameters} \\
         \includegraphics[height=4.5cm]{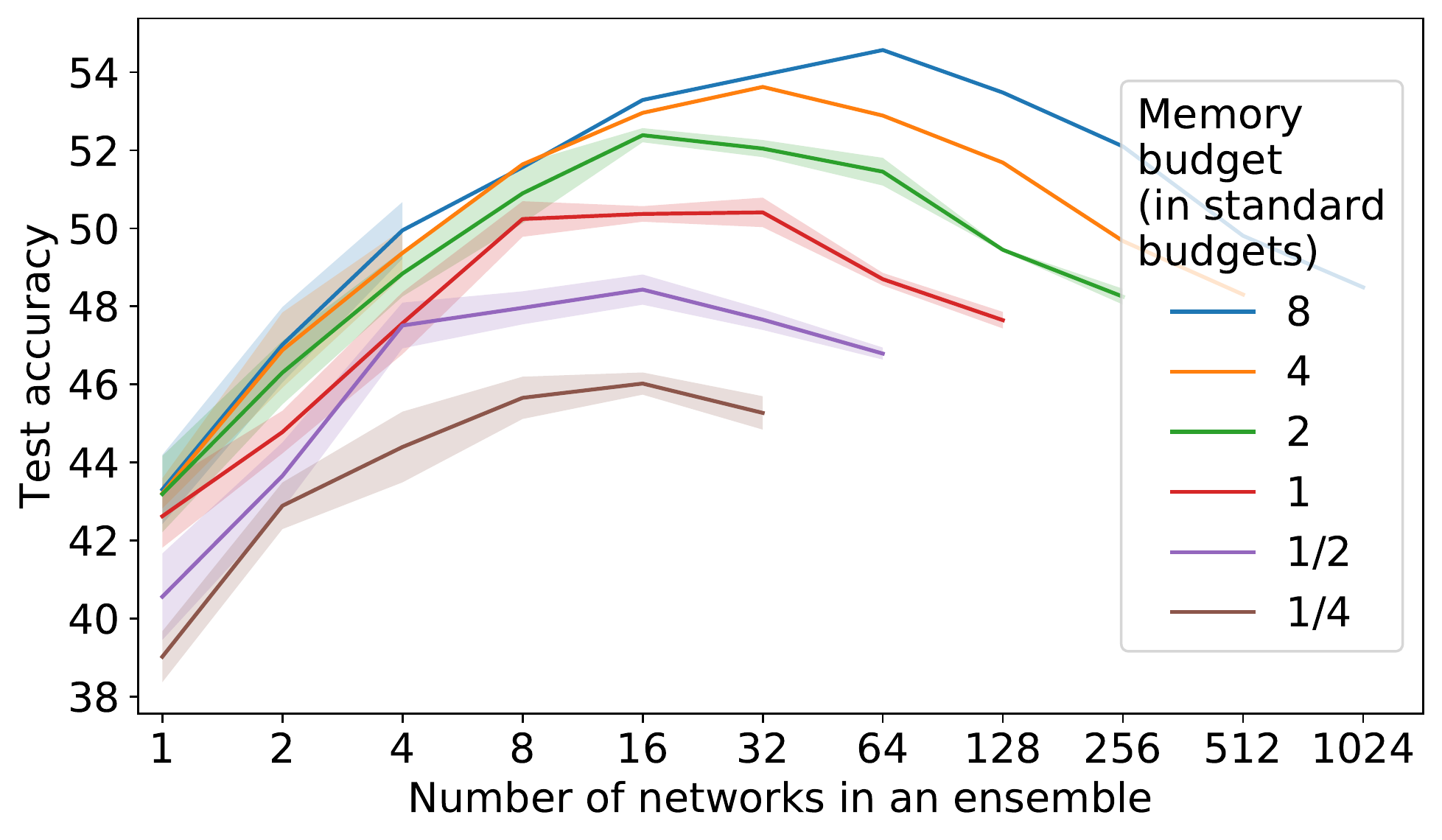}
           & \includegraphics[height=4.5cm]{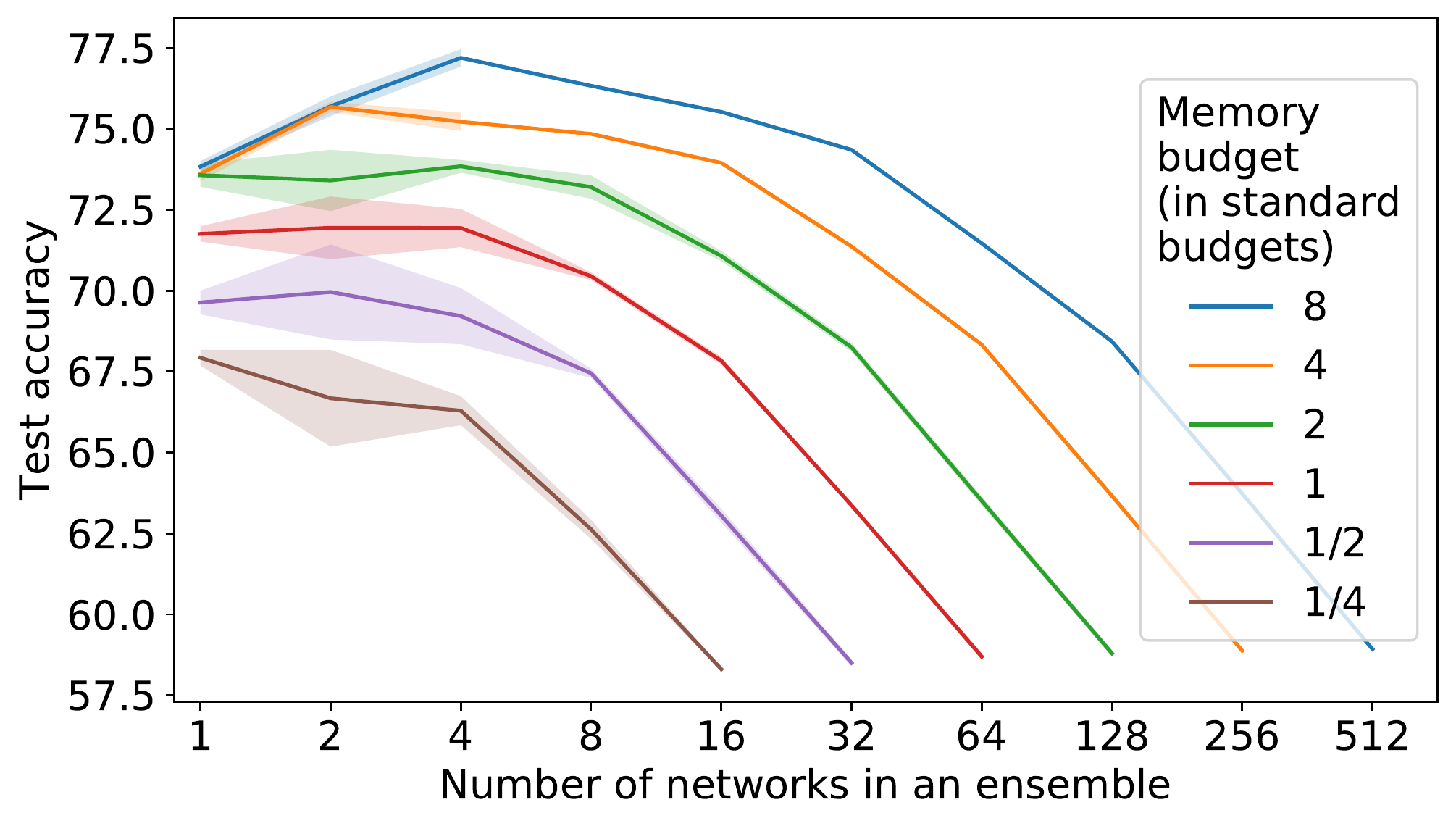} \\
           \small{WideResNet, CIFAR-10, no hyperparameter tuning} & \small{WideResNet, CIFAR-10, with tuned hyperparameters} \\
        \includegraphics[height=4.5cm]{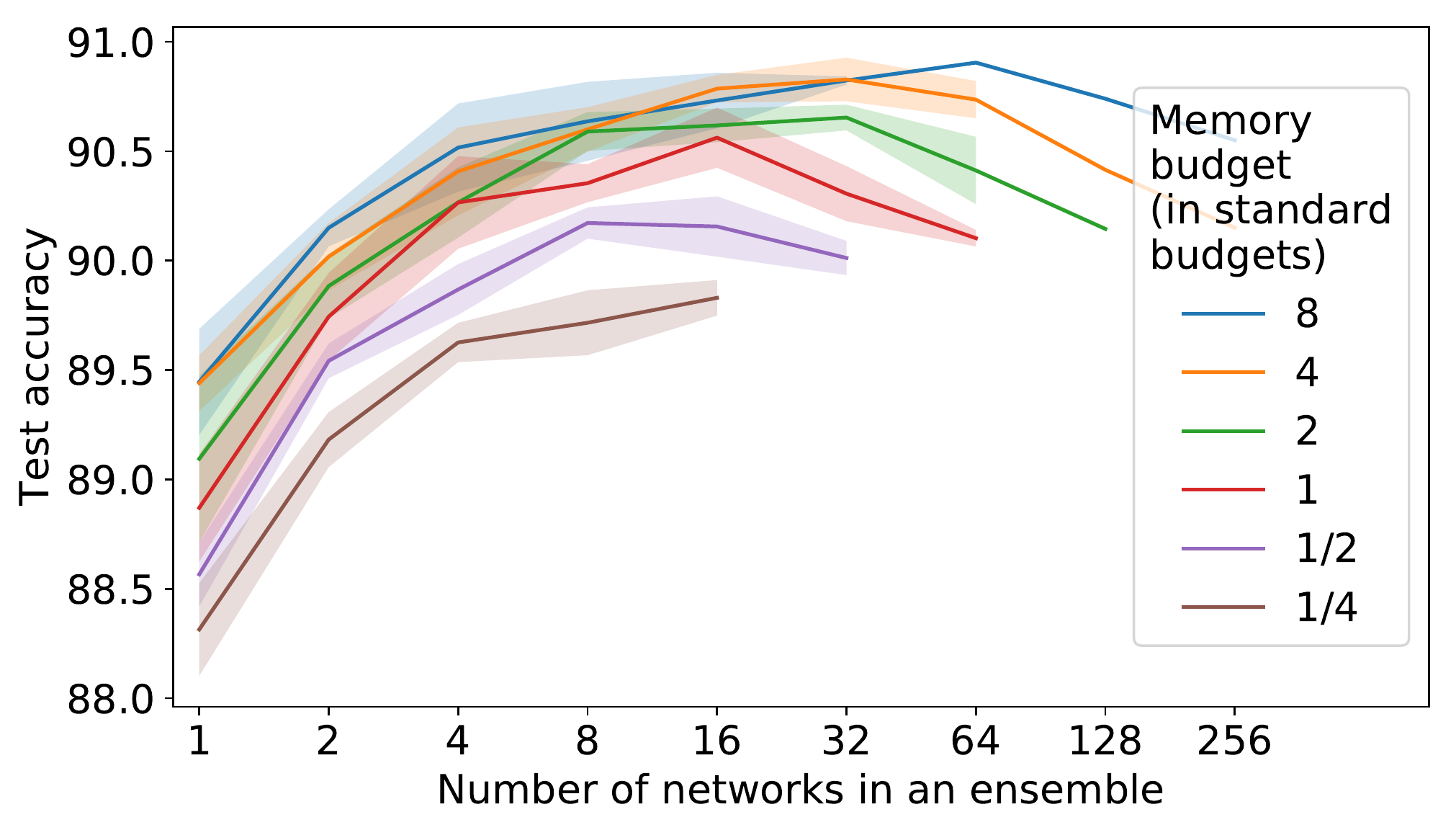} &
           \includegraphics[height=4.5cm]{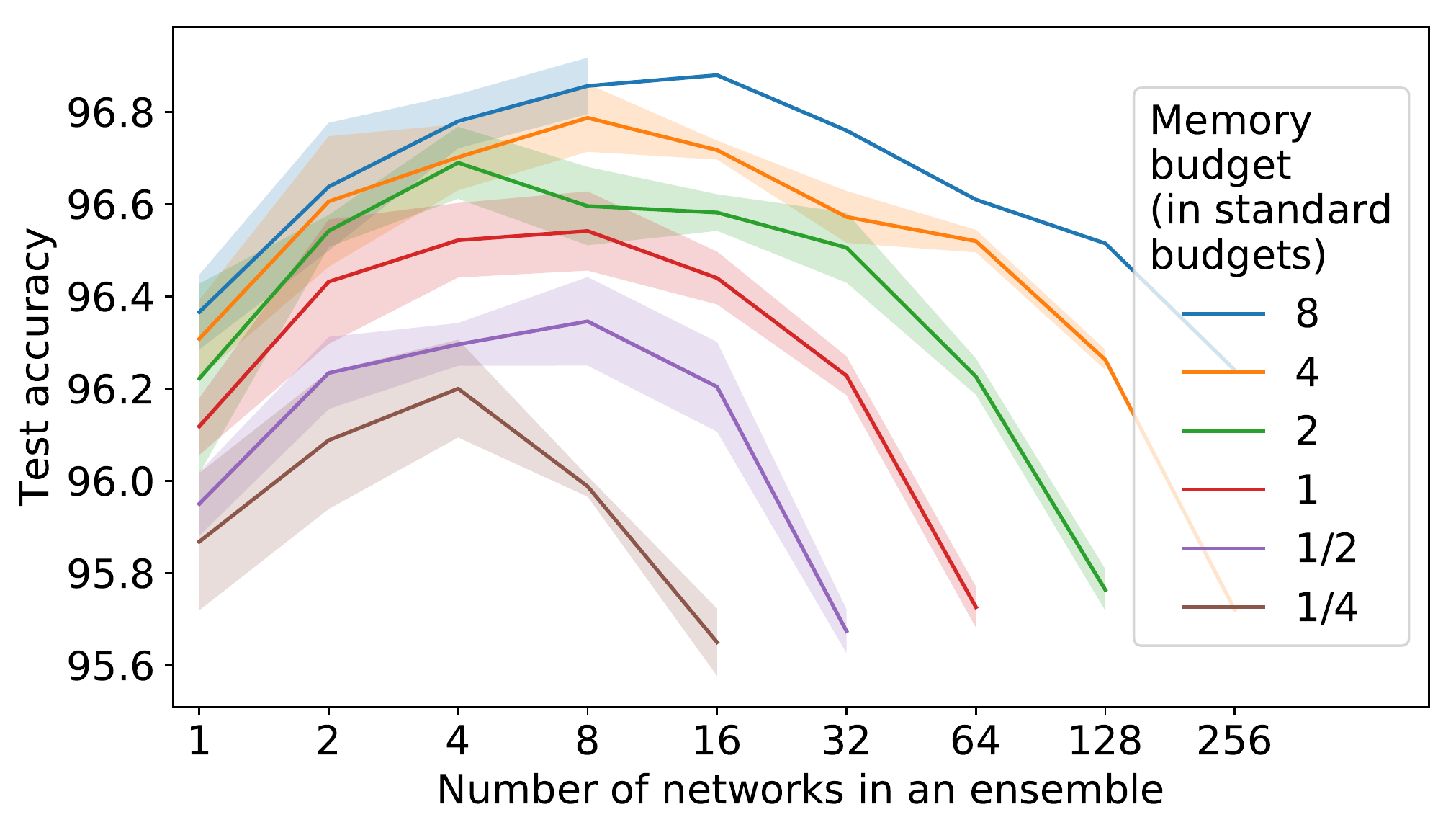}\\
           \small{VGG, CIFAR-10, no hyperparameter tuning} & \small{VGG, CIFAR-10, with tuned hyperparameters} \\
        \includegraphics[height=4.5cm]{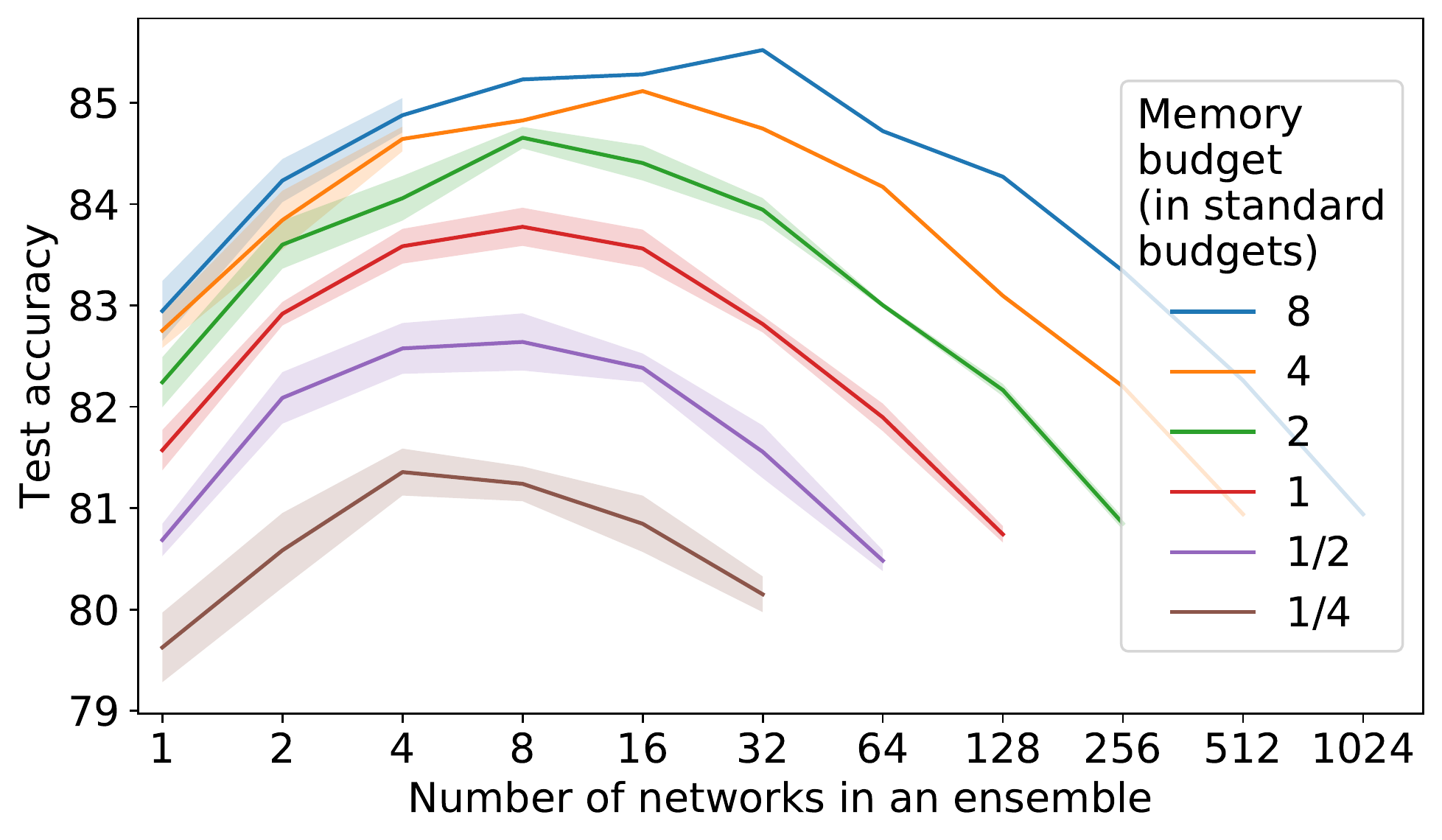} &
           \includegraphics[height=4.5cm]{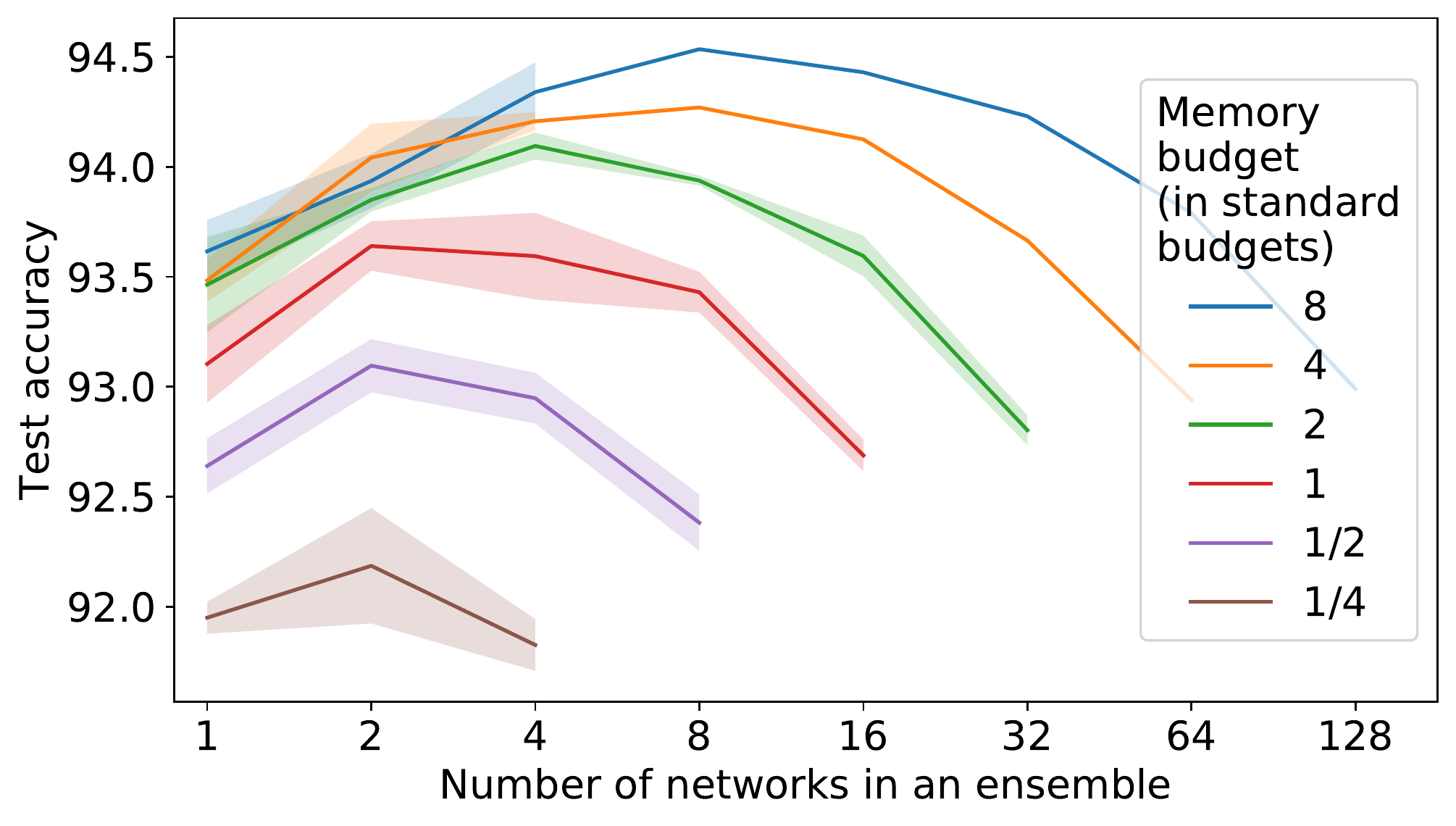}\\
        \end{tabular}
}
\caption{MSA effect for CNNs: mean $\pm$ standard deviation of the test accuracy of the ensembles with fixed memory budget. The x-axis denotes the number of networks $N$ in the ensemble. Each line corresponds to a fixed memory budget (the number of parameters) measured in standard budgets. One standard budget corresponds to the commonly used model size. For some large ensembles, the standard deviation is not provided due to the limited computational resources.}
\label{fig:msa_cnns}
\end{center}
\vskip -0.2in
\end{figure*}

\begin{figure*}[h!]
\vskip 0.05in
\begin{center}
\centerline{
 \begin{tabular}{cc}
        \small{Transformer, IWSLT De-En, no hyperparameter tuning} & \small{Transformer, IWSLT De-En, with tuned hyperparameters} \\
        \includegraphics[height=4.5cm]{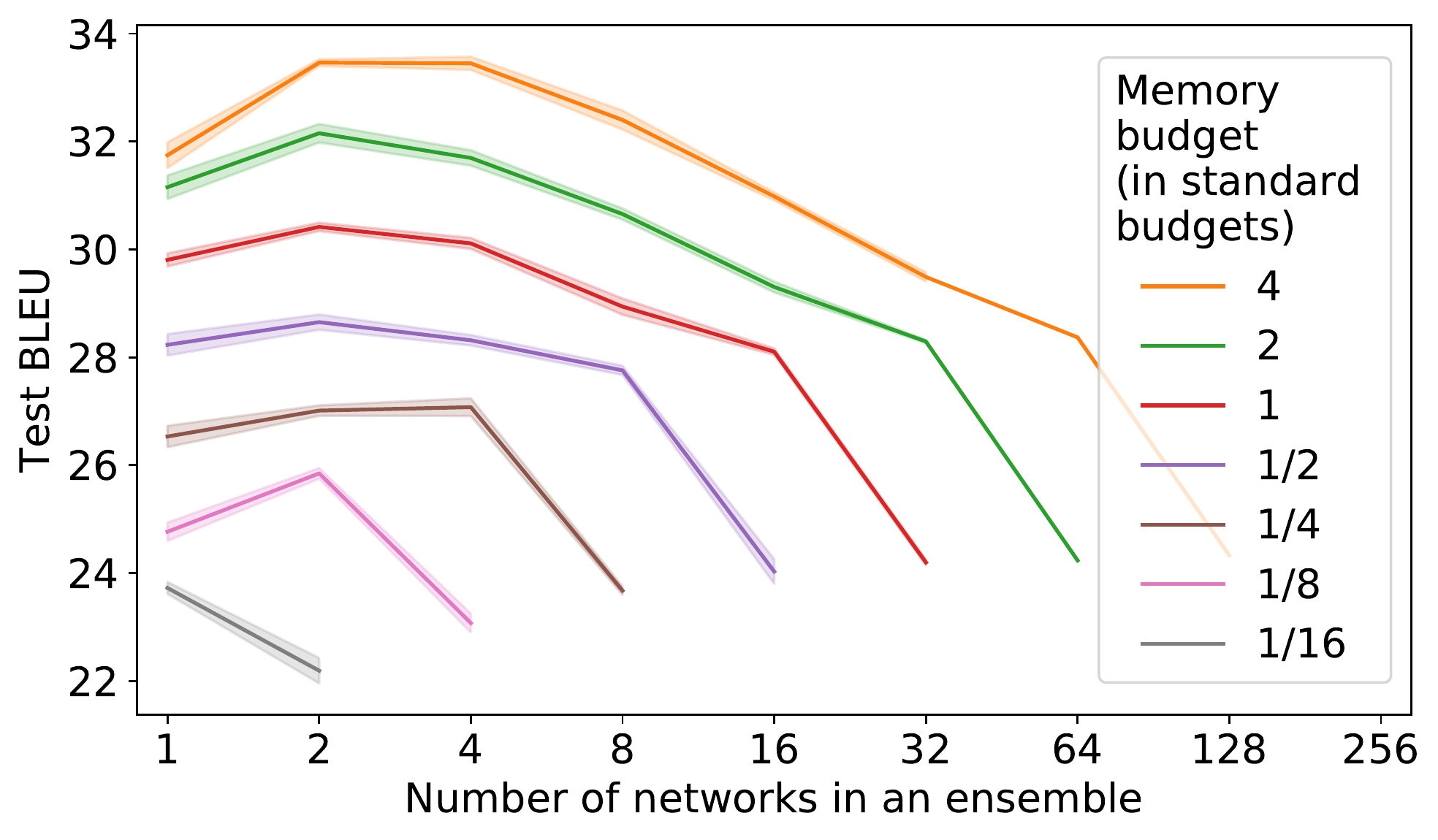}
           & \includegraphics[height=4.5cm]{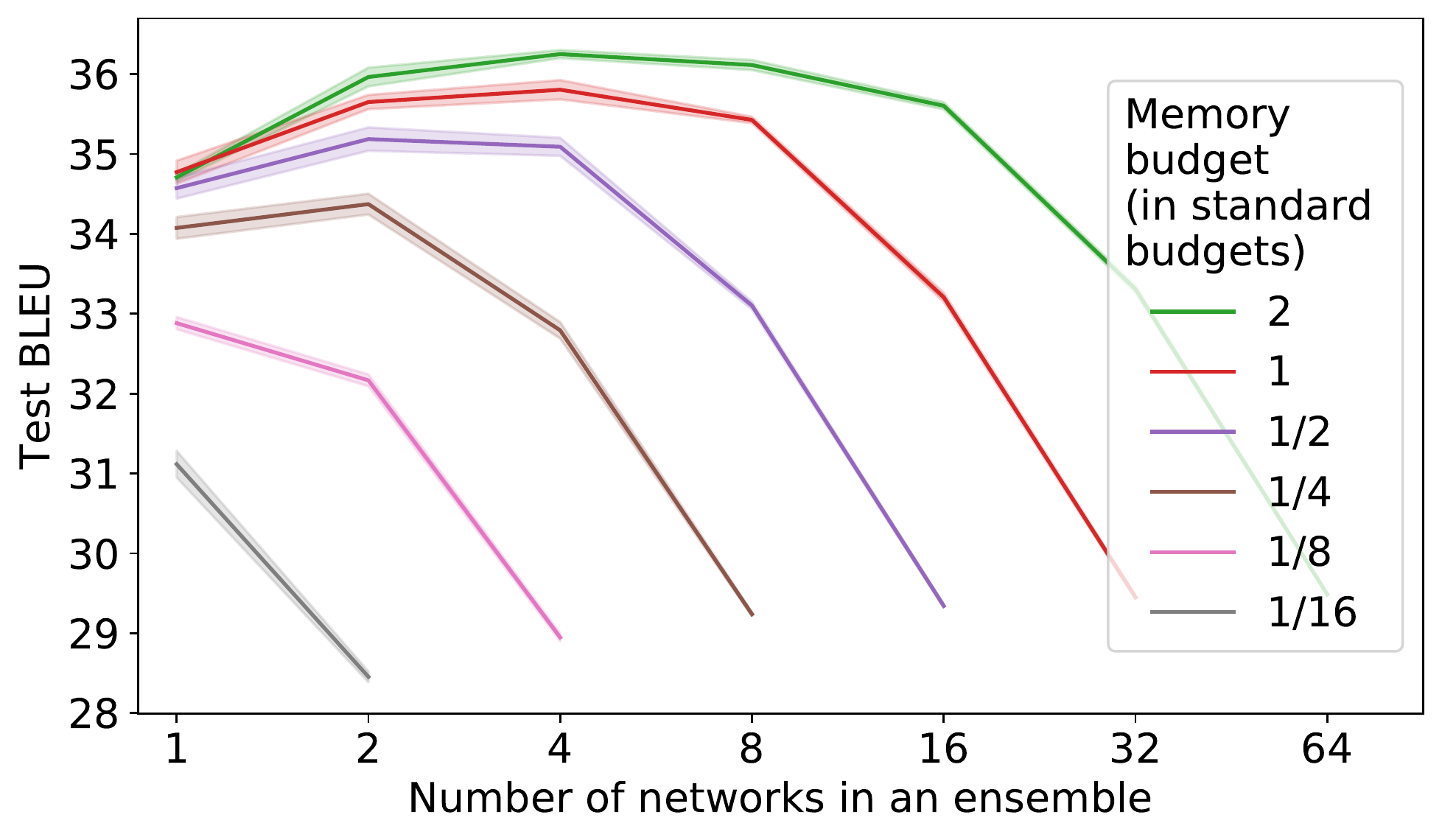}
        \end{tabular}
}
\caption{MSA effect for Transformer: mean $\pm$ standard deviation of BLEU of the ensembles with fixed memory budget. The x-axis denotes the number of networks $N$ in the ensemble. Each line corresponds to a fixed memory budget (the number of parameters) measured in standard budgets. One standard budget corresponds to the size of one Transformer with $k=512$. For some large ensembles, the standard deviation is not provided due to the limited computational resources.}
\label{fig:msa_transformer}
\end{center}
\vskip -0.2in
\end{figure*}

The memory split plots for CNNs and Transformer are given in figures~\ref{fig:msa_cnns} and~\ref{fig:msa_transformer} correspondingly. Two columns show the results for two settings (without hyperparameter tuning and with tuned hyperparameters), and each row corresponds to a dataset--architecture pair. 

\paragraph{Verifying assumptions for the MSA effect.}
Our experiments confirm the generally accepted view that increasing the size $S$ of a single model results in an increase and saturation of test quality $Q(E(1, S))$, with and without hyperparameter tuning for each model size. We refer to this effect as individual quality saturation. This individual quality saturation may be seen from figures~\ref{fig:msa_cnns} and \ref{fig:msa_transformer}: going from bottom to top along the vertical line at abscissa $N=1$ (moving among colored lines), but we also give the more convenient plots in Appendix~\ref{app:width}. The similar results without hyperparameter tuning, and so with mere regularization, were shown in~\cite{openai}, but we did not find a neat similar experiment for a regularized setting. 

We also checked that, when the member network size $S$ is fixed, increasing the ensemble size $N$ results in an increase and saturation of $Q(E(N, S))$ (referred later as ensemble quality saturation). This observation may also be made from figures~\ref{fig:msa_cnns} and \ref{fig:msa_transformer}, and the similar results are given in~\cite{pitfalls}. Two described observations allow us to move ahead and answer the research questions stated in section~\ref{sec:motiv}.

\paragraph{RQ1: Does the MSA effect hold for various datasets and architectures?}
For a particular memory budget $B$, the MSA effect holds, if the number of networks in the optimal memory split is larger than one:
\[
N_* > 1, \quad N_* = \underset{N \geqslant 1}{\text{argmax}} \, Q (E(N, B / N)).
\]
In other words, the line on the plot, corresponding to budget $B$, has an optimum at abscissa $N > 1$. We observe the MSA effect for all considered dataset--architecture pairs, for a wide range of budgets. For example, consider WideResNet with tuned hyperparameters on CIFAR-100, and a memory budget, equivalent to one standard model --- WideResNet-28-10. The red line, corresponding to this budget, has an optimum at abscissa $N=16$. That is, sixteen WideResNets of the width factor $4\times$ smaller than the standard one\footnote{The number of parameters in a network is a quadratic function of the width factor.}, perform significantly better than one standard WideResNet-28-10 (82.52\% test accuracy v.s. 80.60\%).

The MSA effect holds for both designed settings. The results in setting A are more smooth in the sense that there is no additional noise caused by hyperparameter tuning, but quality for setting A is low due to the absence of proper regularization. In setting B, 
hyperparameter optimization is approximate,
and for some model sizes, the found hyperparameters may be more suitable, than for others. To make the hyperparameter search fairer, we used the same uniform grid for all model sizes. We perform additional experiments on the hyperparameter search below.

\begin{figure*}[h!]
\vskip 0.05in
\begin{center}
\centerline{
    \begin{tabular}{ccc}
        \includegraphics[height=5.5cm]{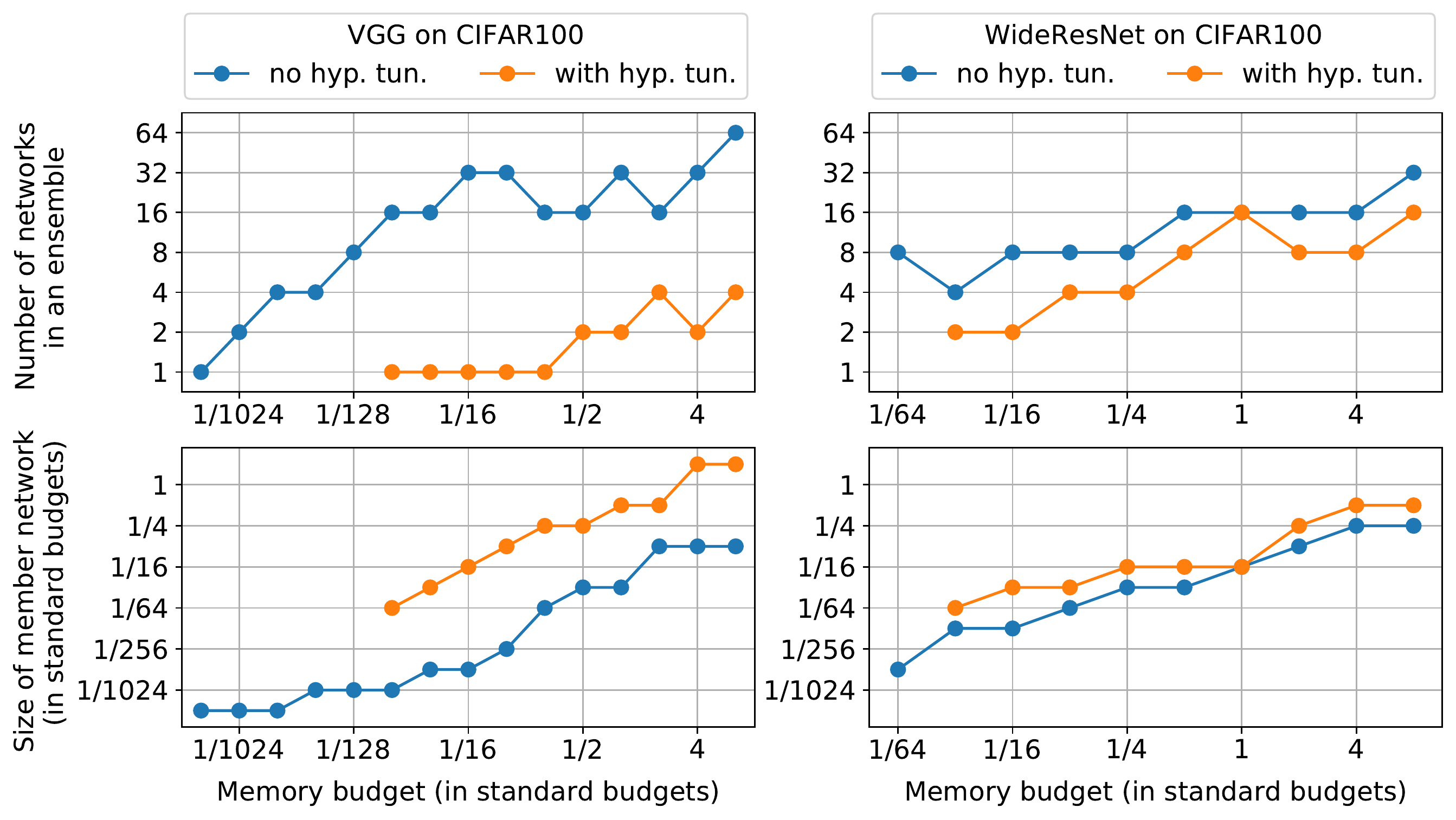} & \hspace{0.3cm} &
        \includegraphics[height=5.5cm]{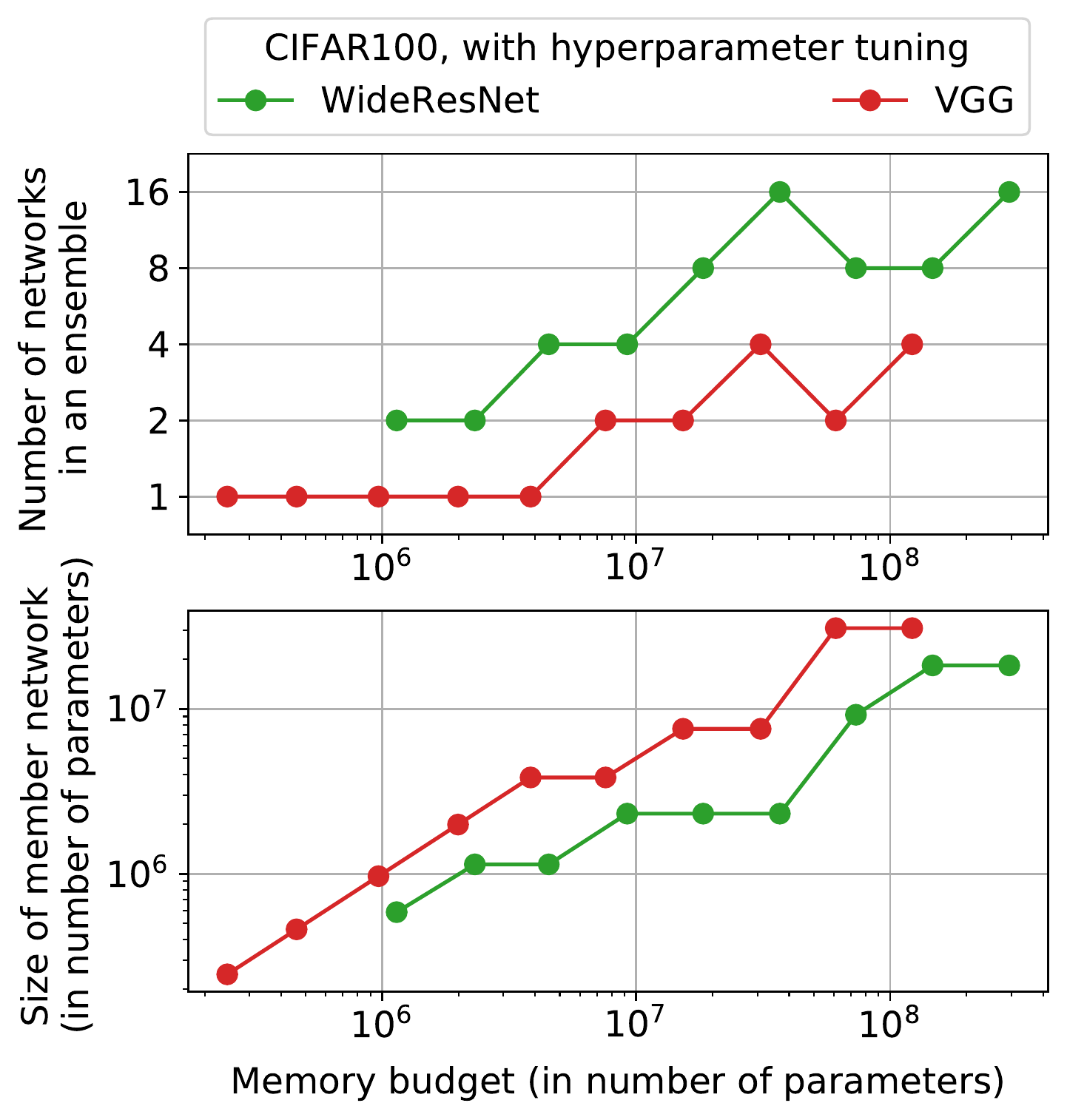}
    \end{tabular}
}
\caption{Optimal memory splits for VGG and WideResNet on CIFAR-100 for different memory budgets. For each optimal memory split the ensemble size is shown in the top plot, while the corresponding size of the member network is shown in the bottom plot. Left: results for settings with and without hyperparameter tuning. Right: comparison of memory splits for VGG and WideResNet, aligned in terms of number of parameters.}
\label{fig:opts}
\end{center}
\vskip -0.2in
\end{figure*}

\paragraph{RQ2: For which budgets does the MSA effect hold?}

The MSA effect holds for all considered budgets, except several smallest ones. 
Notably, for each considered architecture, the MSA effect holds for the standard budget $B_{standard}$, denoted in the red line on all plots.
This means, that widely used configurations of popular architectures are not optimal, and a simple technique, memory splitting, could significantly improve their quality, retaining the same number of parameters. In the majority of cases, the MSA effect also holds for budgets, several times smaller than the standard one.

For large budgets, the MSA effect is expected because of the individual quality saturation: if we split $B=8 B_{standard}$, $E(2, 4 B_{standard})$ easily outperforms $E(1, 8 B_{standard})$ because quality gap between member networks $E(1, 8 B_{standard})$ and $E(1, 4 B_{standard})$ is negligible. However, for smaller budgets, there is usually a significant quality gap between the single network and the member network of the optimal memory split. Ensembling member networks pushes quality of the memory split up enough to cause the MSA effect.

\paragraph{RQ3: How does an optimal split look like for different budgets?}

Figure~\ref{fig:opts} shows optimal memory splits for VGG and WideResNet on CIFAR-100 for different memory budgets. The results on other dataset--architecture pairs look similar and are presented in Appendix~\ref{app:opt_split}. We would like to note that the depicted values of optimal ensemble size and member network size are approximate, due to the  discreteness of the ensemble size grid, and the variance in accuracy of neural networks, trained from different random seeds.

For all dataset--architecture pairs, with increasing memory budget, the optimal memory split grows both in terms of ensemble size and member network size. Hence, there is no one global optimal ensemble size or member network size. For small budgets, the optimal decision is not to split memory and therefore the optimal ensemble size does not grow (see small budgets for VGG with hyperparameter tuning). For large budgets, the quality of one network saturates, therefore the optimal member network size saturates too (see large budgets for all tasks).

In the case of VGG and WideResNet, hyperparameter tuning mostly consists of choosing the right regularization and therefore, first of all, makes large networks better. As a result, optimal memory splits for the setting with hyperparameter tuning contains fewer networks of larger size. 

To compare optimal memory splits between different architectures, we depict the results for VGG and WideResNet on CIFAR-100 with hyperparameter tuning together in one plot (figure~\ref{fig:opts}, right). We align the budgets by measuring them in number of parameters. While achieving much higher quality than VGG, WideResNet also uses parameters more efficiently inside one network. Hence, for WideResNet, the optimal memory split  consists of smaller networks, and the optimal ensemble size becomes greater than one, starting from much smaller budgets, than for VGG. 

\paragraph{Tuning hyperparameters with Bayesian optimization.}
\label{BO}
In order to better justify that the MSA effect holds in the regularized setting, we conduct additional experiments with more careful hyperparameter tuning. We use Bayesian optimization (BO)~\cite{bayesopt} to find hyperparameters better than ones selected by grid search.

For different model sizes of WideResNet on CIFAR-100, we choose the best learning rate, weight decay and dropout rate on the validation set using BO with 20 iterations. We use an open-source implementation\footnote{ \url{https://github.com/fmfn/BayesianOptimization}} of BO. BO takes more time than grid search, so we omit largest budgets in this experiment.

\begin{figure}[h!]
\vskip 0.1in
\begin{center}
\centerline{
        \begin{tabular}{c}
        \small{WideResNet, CIFAR-100, BO-tuned hyperparameters} \\
        \includegraphics[height=4.35cm]{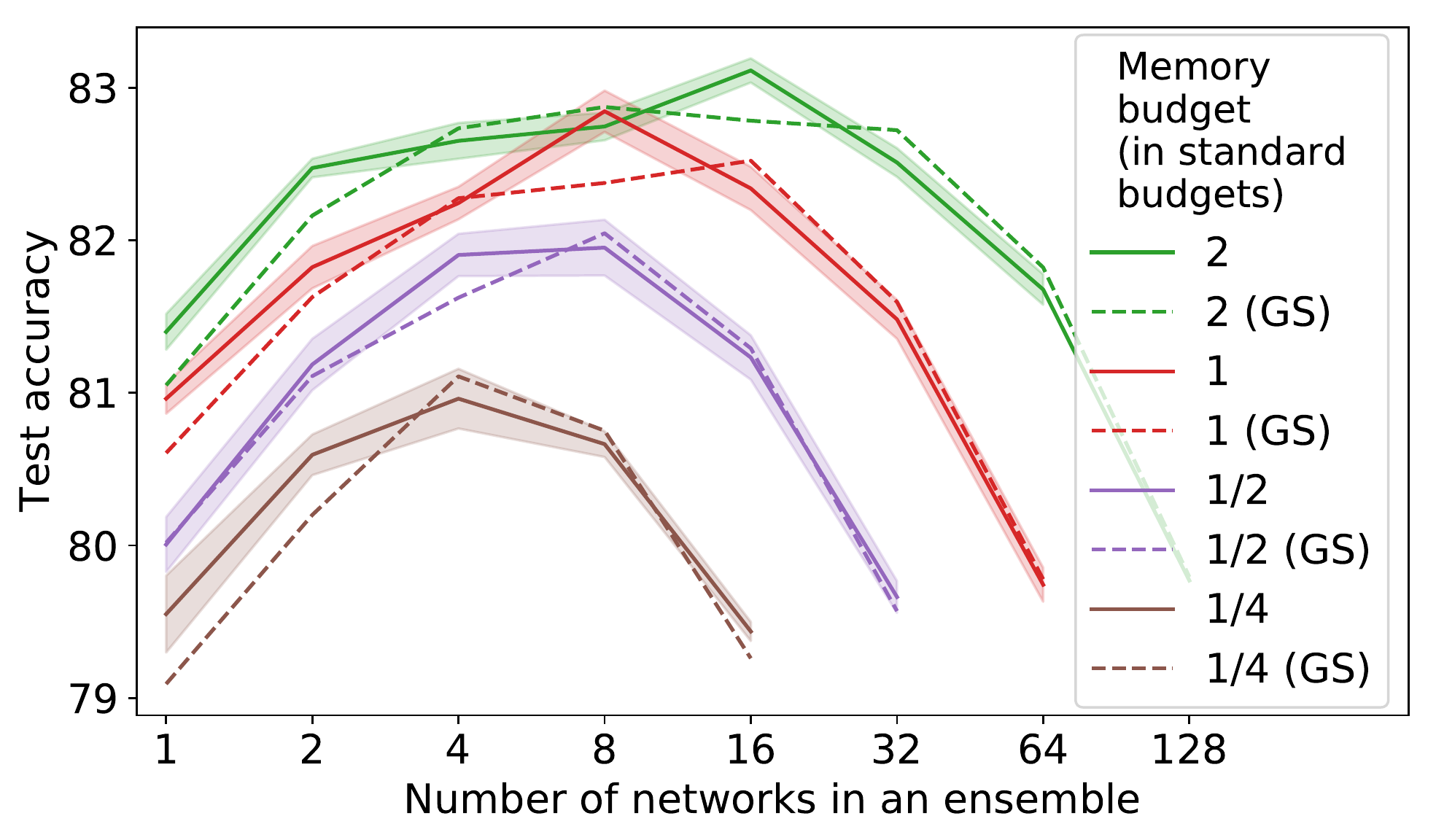}
        \end{tabular}
}
\caption{MSA effect for setting with hyperparameters tuned using Bayesian optimization, WideResNet on CIFAR-100. Each solid line shows mean $\pm$ standard deviation of the test accuracy of the ensembles with a fixed memory budget; the x-axis denotes the number of networks $N$ in the ensemble. One standard budget corresponds to the size of one WideResNet-28-10. The dashed line corresponds to the results with grid search, for reference.}
\label{fig:bo}
\end{center}
\vskip -0.1in
\end{figure}

The results are given in figure~\ref{fig:bo}. The MSA-effect holds for both hyperparameter tuning methods, while the optimization methods differ considerably: grid search explores the hyperparameter space regularly; and BO incorporates random search, and searching within narrow region near the best discovered points. 

Hyperparameters, found by BO, in most cases result in higher test accuracies of single models, than hyperparameters found by grid search. This is visible at abscissa $N=1$ in figure~\ref{fig:bo}. However, the test accuracies of memory splits are not always higher for BO than for grid search. This is because optimal hyperparameters for the ensemble may differ from optimal hyperparameters for the single model. Optimizing hyperparameters for ensemble is extremely expensive, but potentially it could make memory splits even stronger, compared to single models. In total, the maximum achieved test accuracy, for the largest considered budget, is higher for BO (82.75\%), than for grid search (82.52\%).

\begin{figure}[h!]
\begin{center}
\centerline{
    \begin{tabular}{c}
        \small{WideResNet, CIFAR-100, tuned hyperparameters} \\
        \includegraphics[height=4.35cm]{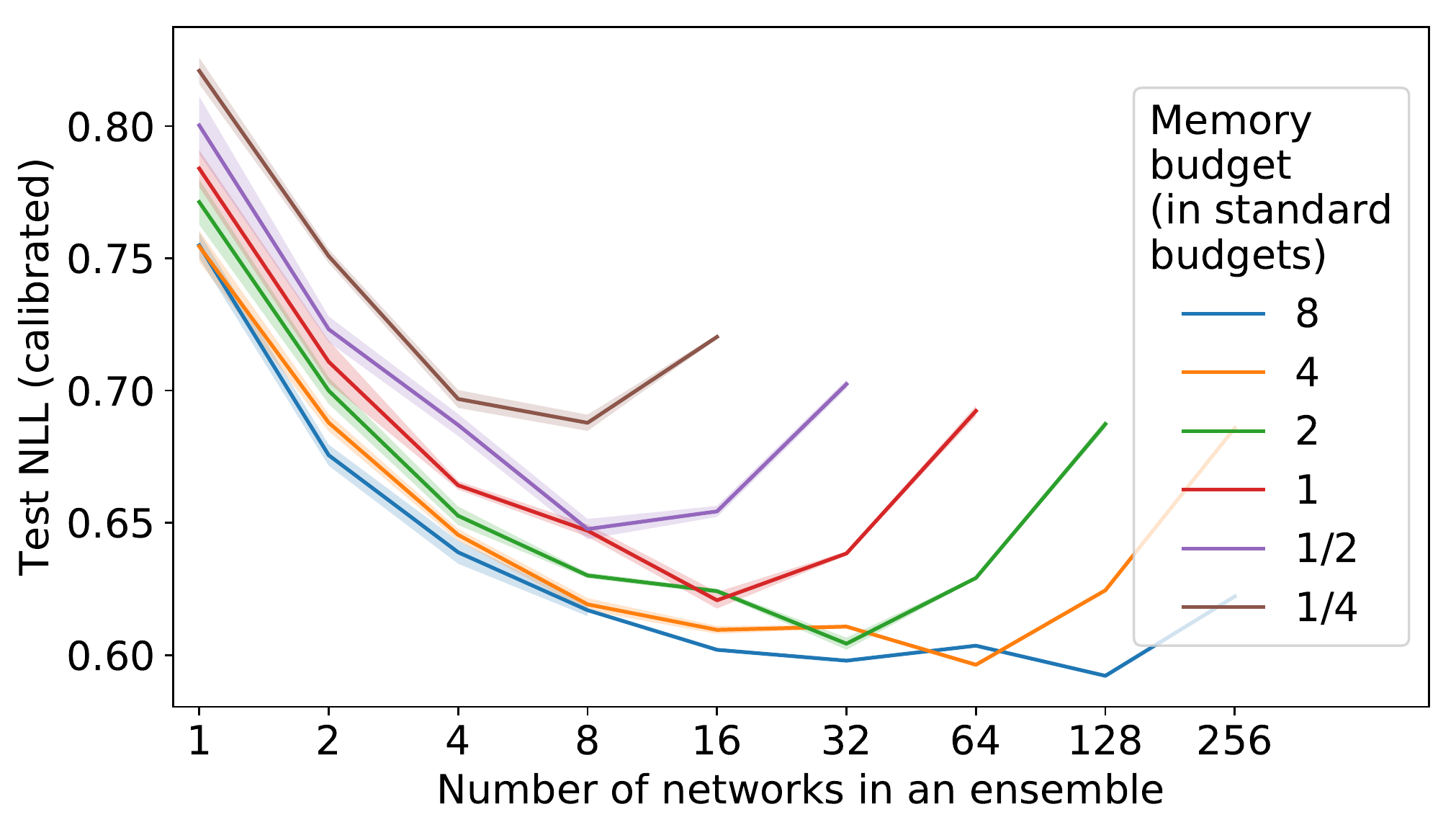}\\
        \small{WideResNet, CIFAR-10, tuned hyperparameters} \\
        \includegraphics[height=4.35cm]{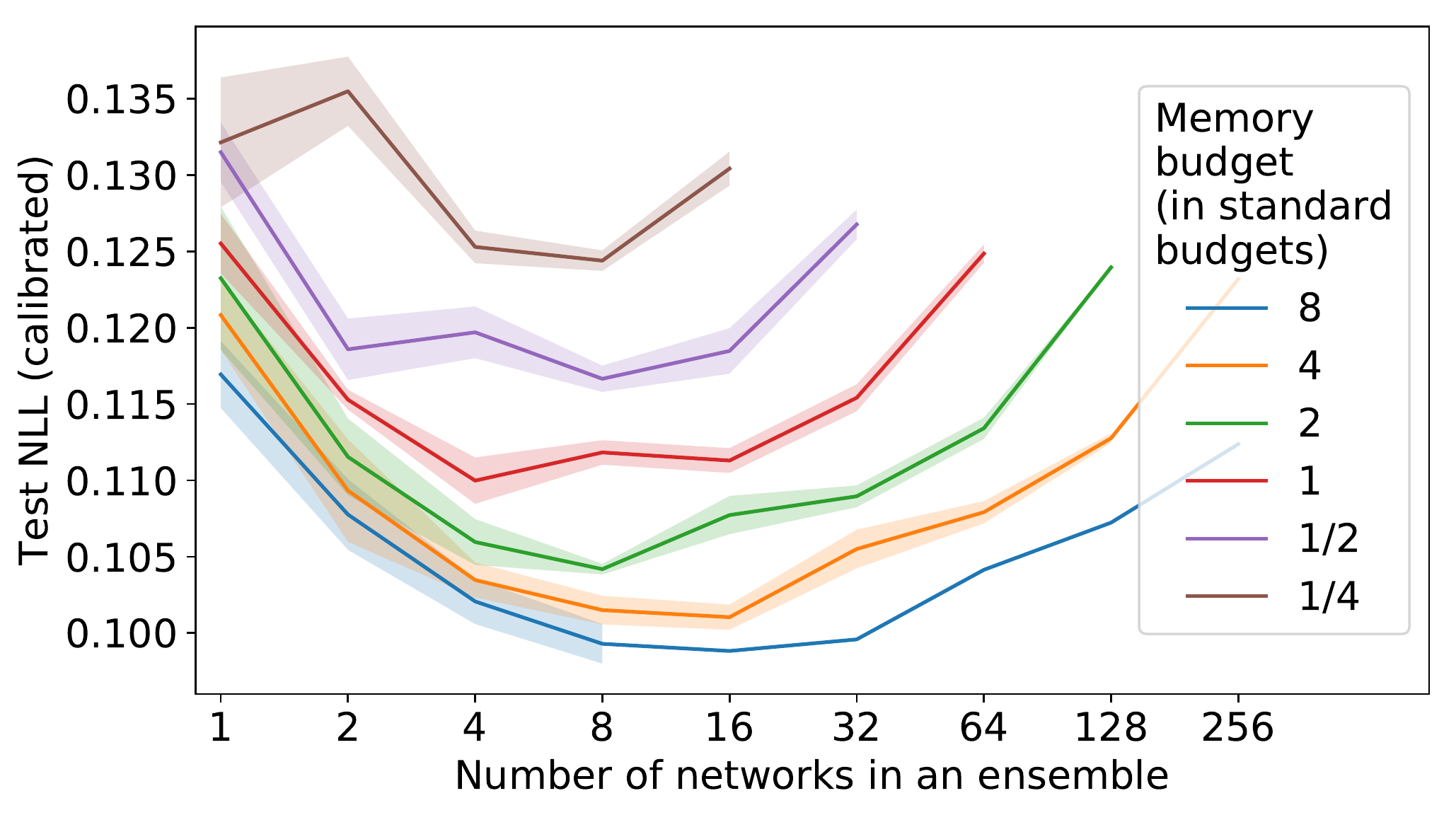}
    \end{tabular}
}
\caption{MSA effect for calibrated test NLL, WideResNet on CIFAR-100 and CIFAR-10. Each line shows mean $\pm$ standard deviation of the calibrated test NLL of the ensembles with a fixed memory budget; the x-axis denotes the number of networks $N$ in the ensemble. One standard budget corresponds to the size of one WideResNet-28-10. For some large ensembles, the standard deviation is not provided due to the limited computational resources.}
\label{fig:uncertainty}
\end{center}
\vskip -0.2in
\end{figure}

\paragraph{MSA effect for uncertainty estimation}
Ensembles of deep neural networks are often used for uncertainty estimation: probability estimates produced by ensembles are more accurate and reliable than ones produced by a single network~\cite{deepens}, in both in-domain~\cite{pitfalls} and out-of-domain~\cite{ovadia} settings.

However, this comparison is usually performed in a fixed width setting, and so with unequal number of parameters. We show that ensembles outperform the single network in in-domain uncertainty estimation, when the total number of parameters in both models is fixed and approximately equal.

We use the calibrated test negative log-likelihood (NLL) to measure the quality of in-domain uncertainty. \citet{pitfalls} highlight the importance of temperature scaling for both single networks and ensembles, since comparison of log-likelihoods without temperature scaling may lead to arbitrary results. We measure the calibrated test NLL for image classification models trained in the main experiments and discussed in paragraphs RQ1--RQ3. We show the results for WideResNet with tuned hyperparameters in figure~\ref{fig:uncertainty}, more results may be found in Appendix~\ref{append:uncertainty}.

The MSA effect holds for the calibrated test NLL in all the same cases (triplets architecture--dataset--budget) as for the test accuracy. The number of networks in a memory split, optimal w.\,r.\,t. calibrated test NLL, is usually equal or a bit larger than that, optimal w.\,r.\,t. test accuracy. 
\section{Conclusion}
In this work, we introduce the MSA effect and the arising simple method of improving the neural network performance in a limited memory setting. Investigating the MSA effect for various datasets, architectures, and budgets, we find that the effect holds even for small configurations of popular architectures and that the larger the configuration, the bigger the ensemble size $N$ and the network size $S$, corresponding to the optimal memory split. Finding the optimal values of $N$ and $S$ without full computation of the memory split plot is an interesting direction for future work.

\section*{Acknowledgments}
We would like to thank Dmitry Molchanov for the valuable feedback. Results for convolutional neural networks were supported by the Russian Science Foundation grant \textnumero 19-71-30020. Results for Transformer were supported by Samsung Research, Samsung Electronics. This research was supported in part through computational resources of HPC facilities at NRU HSE.

\bibliography{example_paper}
\bibliographystyle{icml2020}

\newpage
\appendix

\section{Experimental details}
\label{append:details}
\subsection{Details for CNNs}
{\bf Data.} We conduct experiments on CIFAR-100 and CIFAR-10 datasets, each containing 50000 training and 10000 testing examples. For tuning hyperparameters, we randomly select 5000 training examples as a validation set. After choosing optimal hyperparameters, we retrain the models on the full training dataset.
We use a standard data augmentation scheme following~\cite{fge}: zero-padding with 4 pixels on each side, random cropping to produce
32×32 images, and horizontal mirroring with probability 0.5. In the experiments without hyper parameter tuning, we do not use data augmentation.

{\bf Training. } Following~\citep{fge}, we train all models for 200 epochs using SGD with momentum of $0.9$ and the following learning rate schedule: constant (100 epochs) -- linearly annealing (80 epochs) --  constant (20 epochs). The final learning rate is 100 times smaller than the initial one. We use the batch size of 128.

{\bf Testing. } We measure the test accuracy and the calibrated negative log-likelihood (NLL) of each ensemble or individual model.

{\bf Bayesian optimization.} We use the following ranges for Bayesian optimization: $[10^{-7}, 0.003]$ for weight decay, $[0, 0.5]$ for dropout, and [$10^{-5}$, $0.05/0.1$] for learning rate for VGG/WideResNet correspondingly.

\subsection{Details for Transformers } 
{\bf Data.} 
We conduct experiments on the IWSLT’14 German-English (De-En) and English-German (En-De) translation tasks~\citep{iwslt} that contain 160K training sentences and 7K validation sentences randomly sampled from the training data. We test on the concatenation of tst2010, tst2011, tst2012, tst2013 and dev2010. We preprocess the data using byte pair encoding (BPE)~\citep{bpe}.

\subsection{Details of the memory splitting procedure}
For a budget $B$ (the number of parameters), we train several memory splits $E(N, B/N)$, each contains $N$ networks of size $B/N$, $N=1, 2, 4, 8\dots$.
The number of parameters in a network is a quadratic function of the width factor. To obtain a network of size $B/N$, we solve the quadratic equation w.\,r.\,t. the width factor, and then round the width factor to the nearest integer.
The difference in the number of parameters between the resulting memory split and budget $B$ is negligible, compared to $B$.

To make predictions with an ensemble, we average predictions of the individual networks after softmax, i.\.e.\,average discrete distributions over classes.

\subsection{Computing infrastructure}
Experiments were conducted on NVIDIA Tesla V100 GPU, Tesla P40 GPU and Tesla P100 GPU.

\section{Individual network quality}
\label{app:width}
\begin{figure*}[h!]
\vskip 0.1in
\begin{center}
\centerline{
 \begin{tabular}{cc}
        \small{WideResNet, CIFAR-100, no hyperparameter tuning} & \small{WideResNet, CIFAR-100, with tuned hyperparameters} \\
        \includegraphics[height=3.7cm]{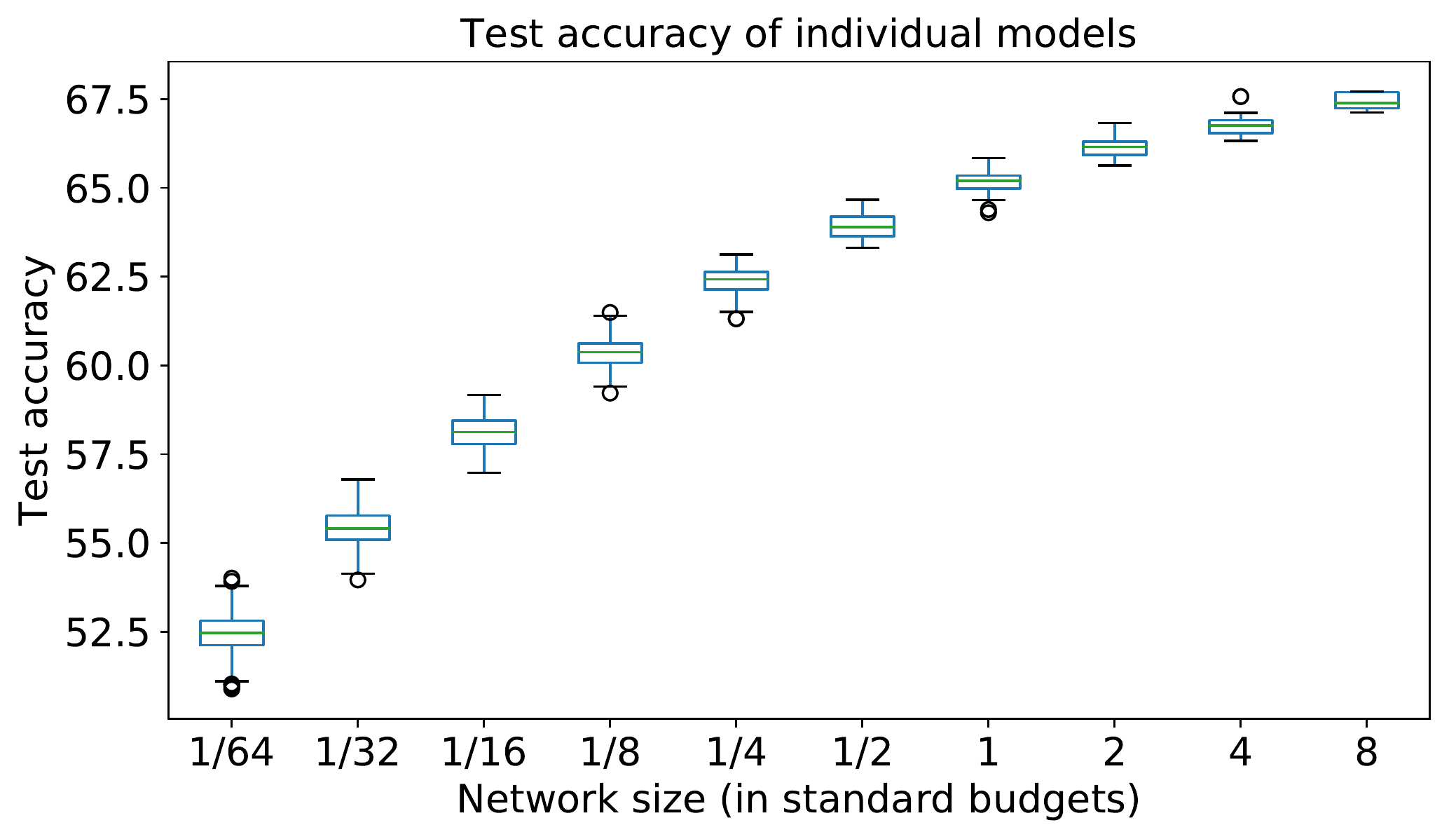}
           & \includegraphics[height=3.7cm]{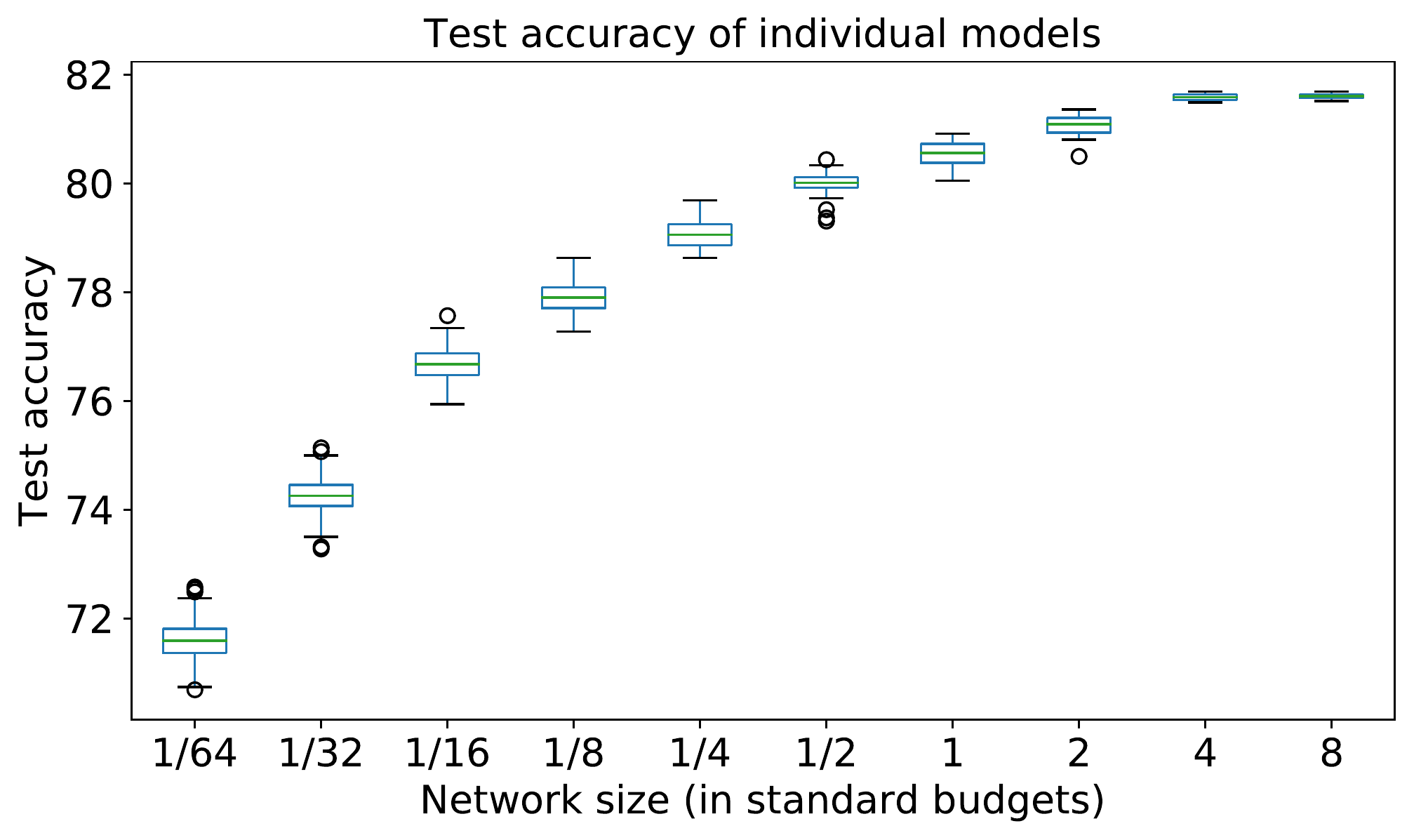} \\
        \small{VGG, CIFAR-100, no hyperparameter tuning} & \small{VGG, CIFAR-100, with tuned hyperparameters} \\
         \includegraphics[height=3.7cm]{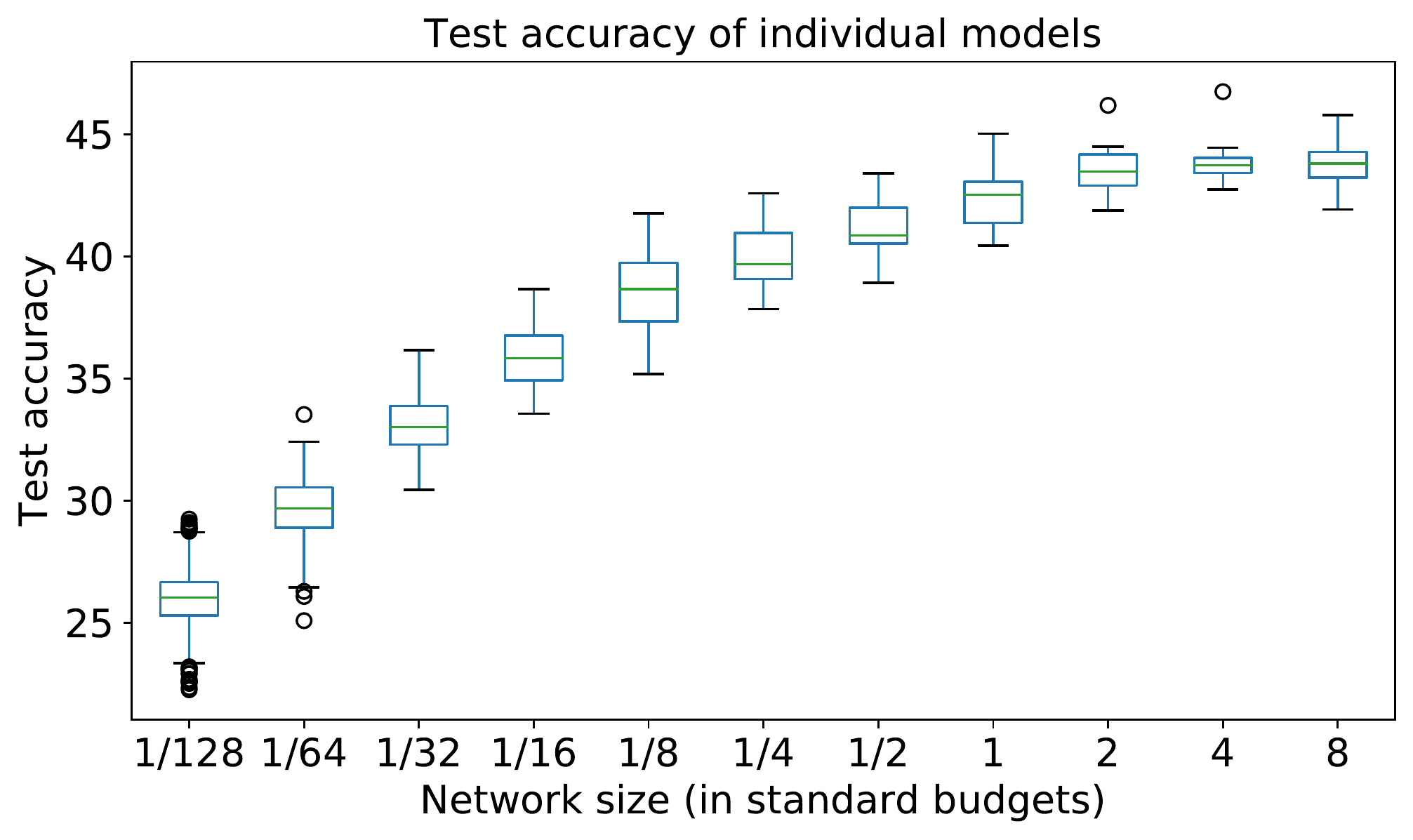}
           & \includegraphics[height=3.7cm]{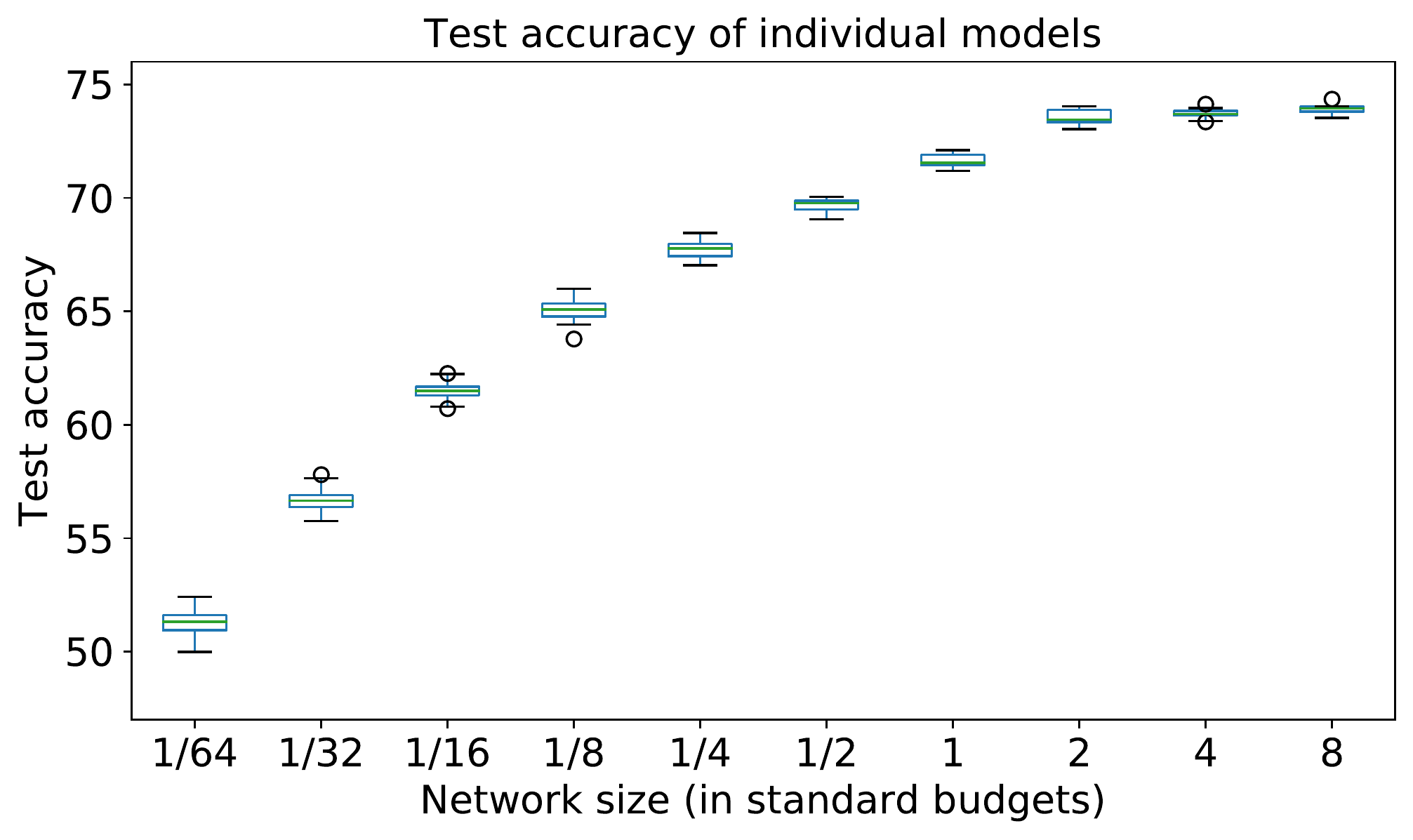} \\
           \small{WideResNet, CIFAR-10, no hyperparameter tuning} & \small{WideResNet, CIFAR-10, with tuned hyperparameters} \\
        \includegraphics[height=3.7cm]{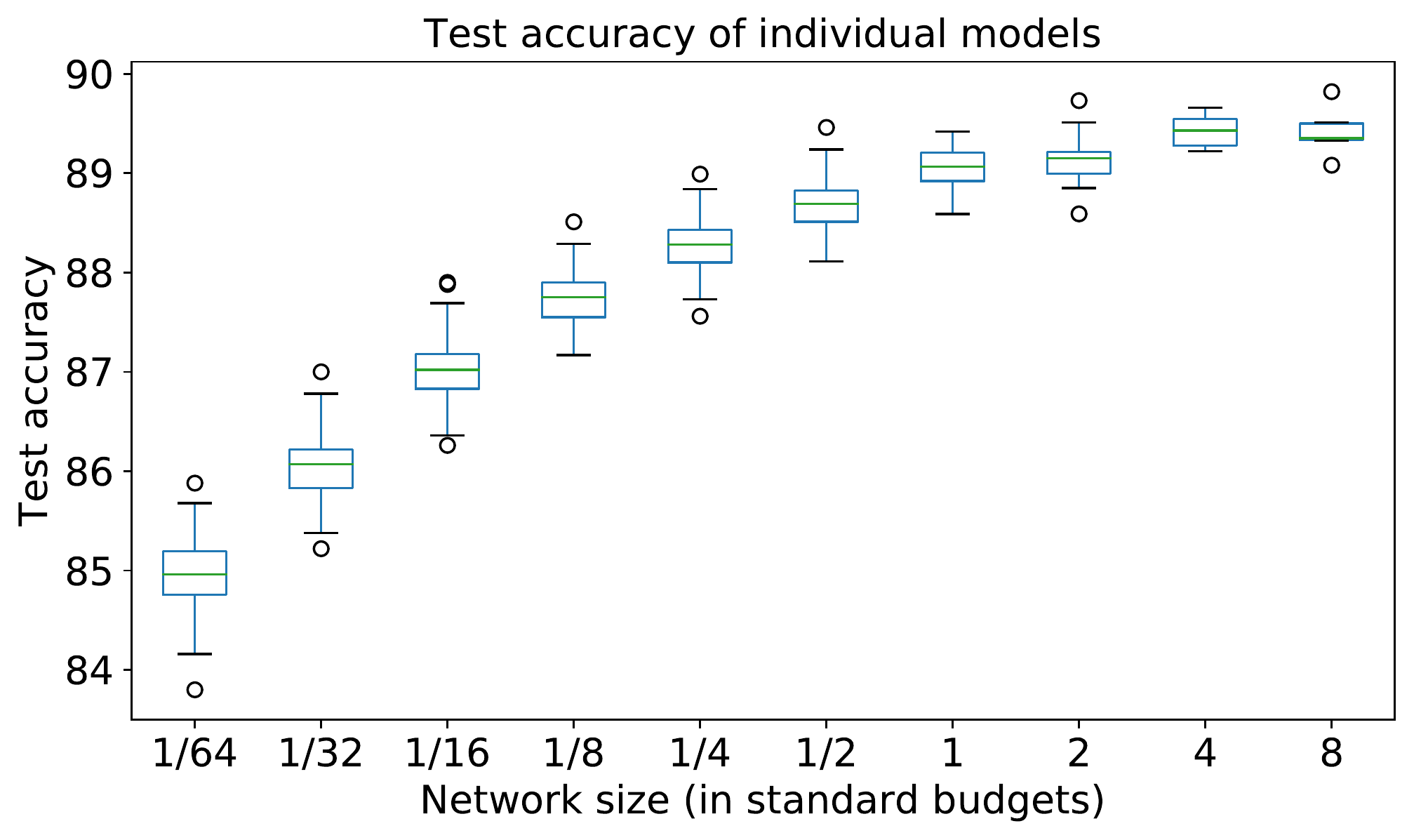} &
           \includegraphics[height=3.7cm]{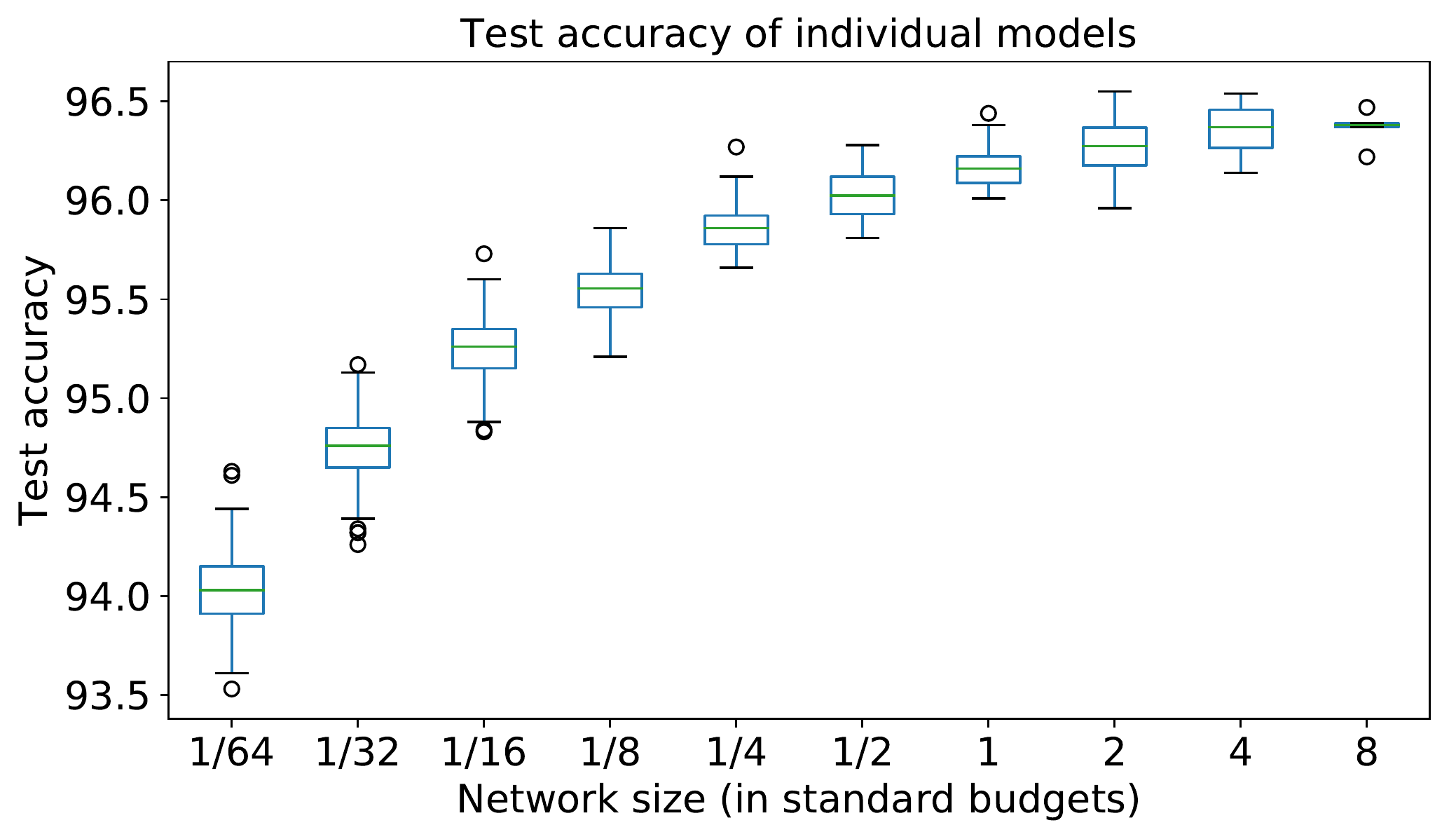}\\
           \small{VGG, CIFAR-10, no hyperparameter tuning} & \small{VGG, CIFAR-10, with tuned hyperparameters} \\
        \includegraphics[height=3.7cm]{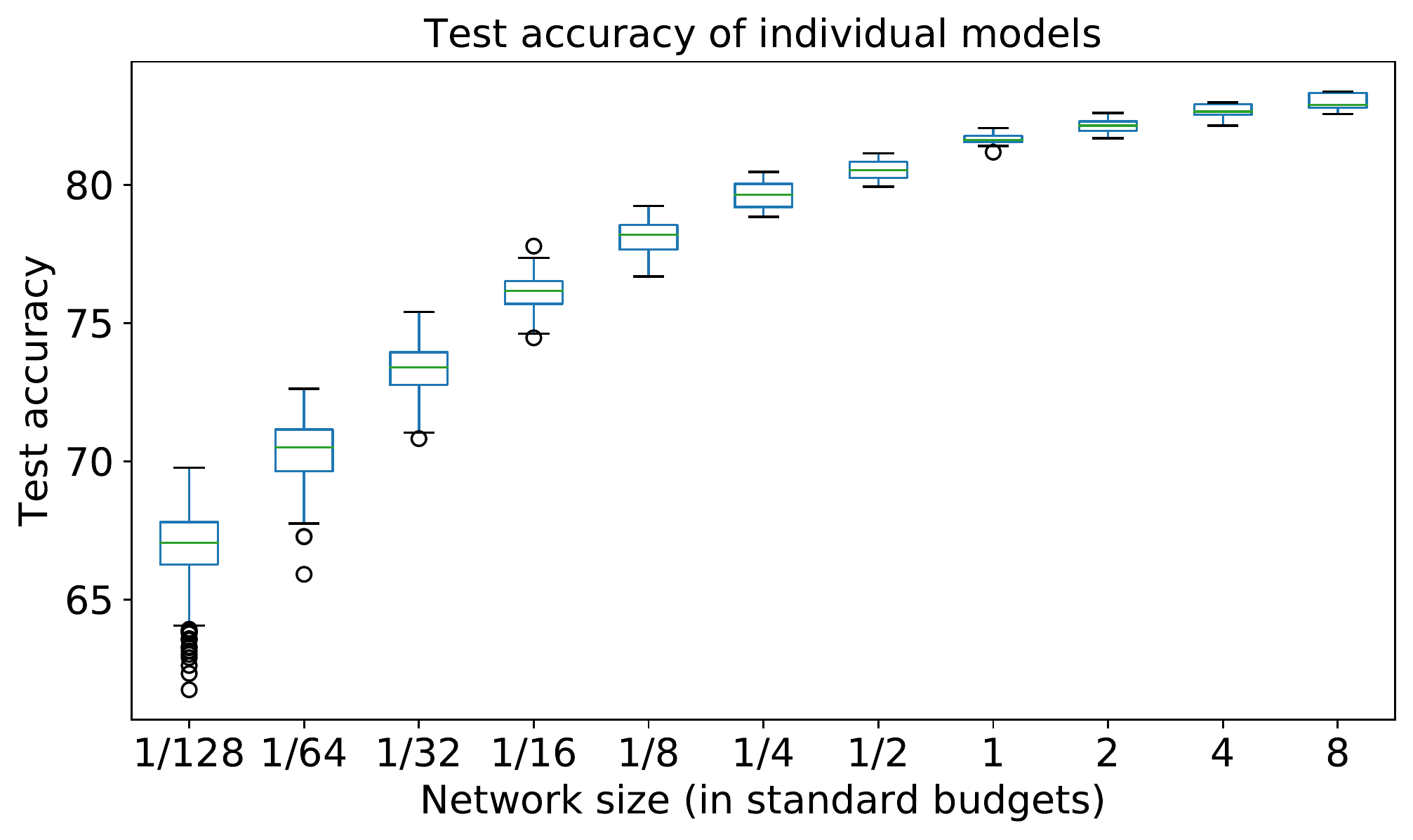} &
           \includegraphics[height=3.7cm]{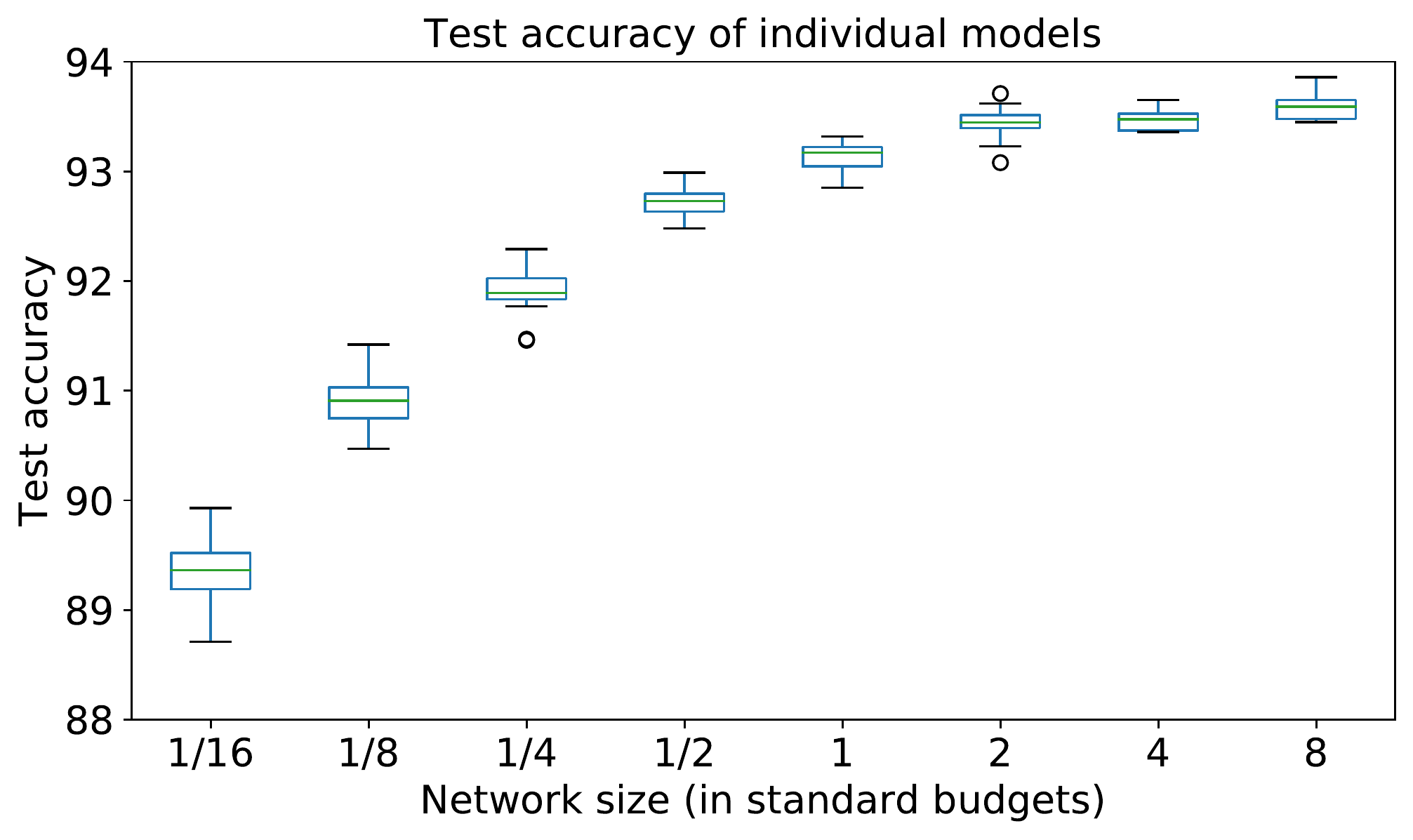}\\
        \small{Transformer, IWSLT De-En, no hyperparameter tuning} & \small{Transformer, IWSLT De-En, with tuned hyperparameters} \\
        \includegraphics[height=3.7cm]{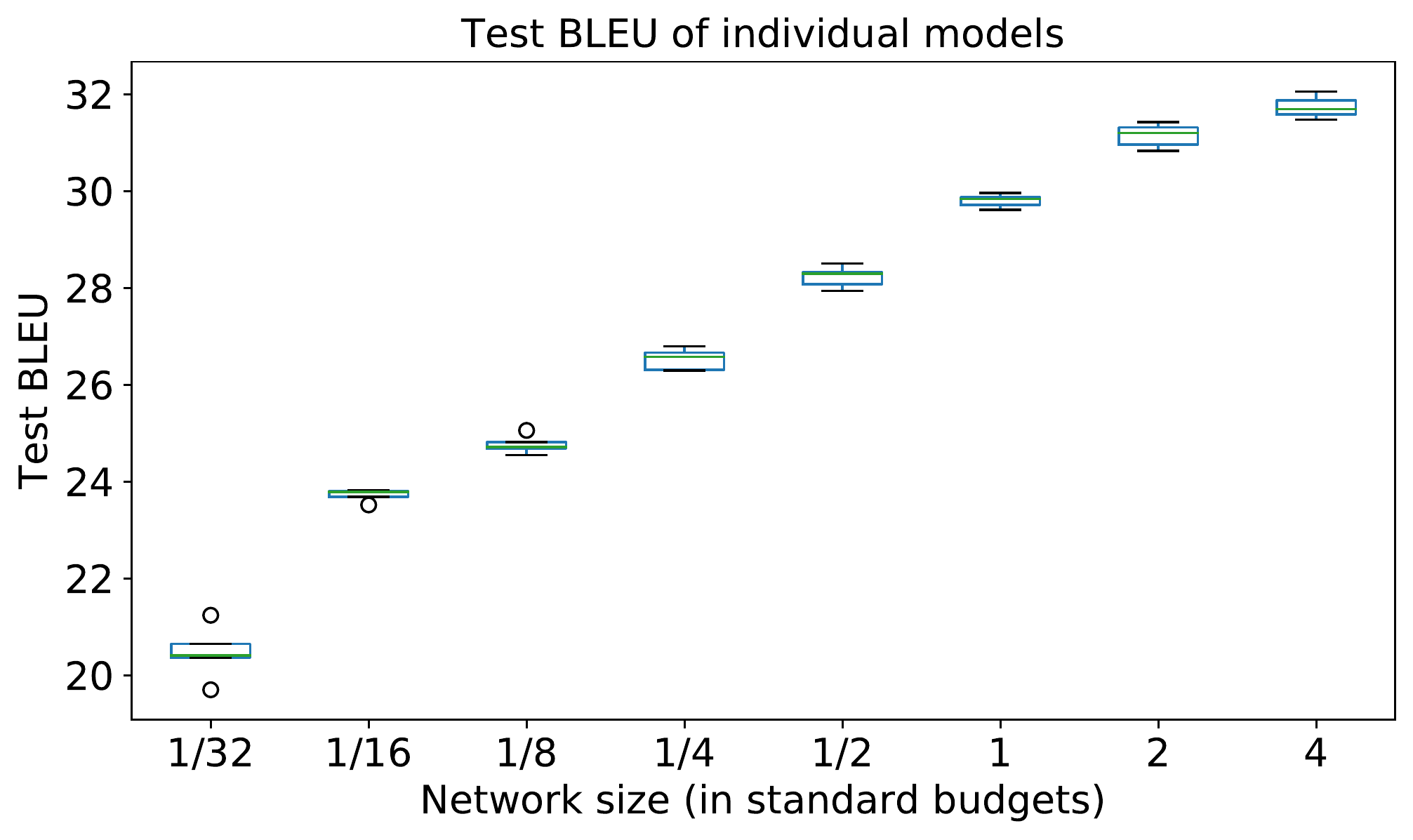}
           & \includegraphics[height=3.7cm]{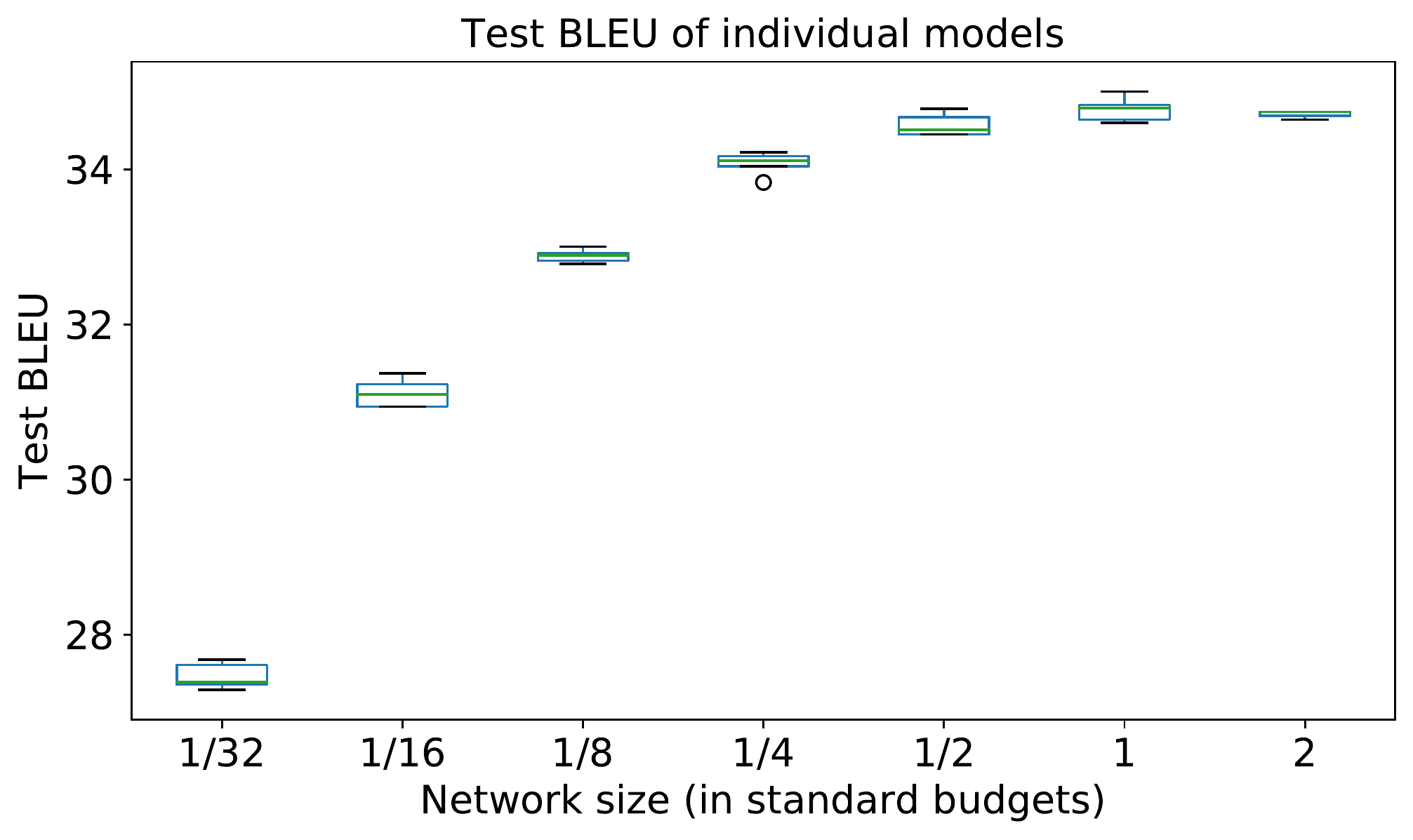} \\
        \end{tabular}
}
\caption{Quality of individual network for different model sizes. To draw boxplots, we use all the networks trained in the memory split experiments.}
\label{fig:acc}
\end{center}
\vskip -0.2in
\end{figure*}

Figure~\ref{fig:acc} shows the test quality of individual network for different model sizes. We present the results for settings with and without hyperparameter tuning for each network size. The results confirm the generally accepted view that increasing the network size leads to the higher quality.

\section{Optimal memory splits}
\label{app:opt_split}
\begin{figure*}[h!]
\vskip 0.05in
\begin{center}
\centerline{
    \begin{tabular}{ccc}
        \includegraphics[height=5.5cm]{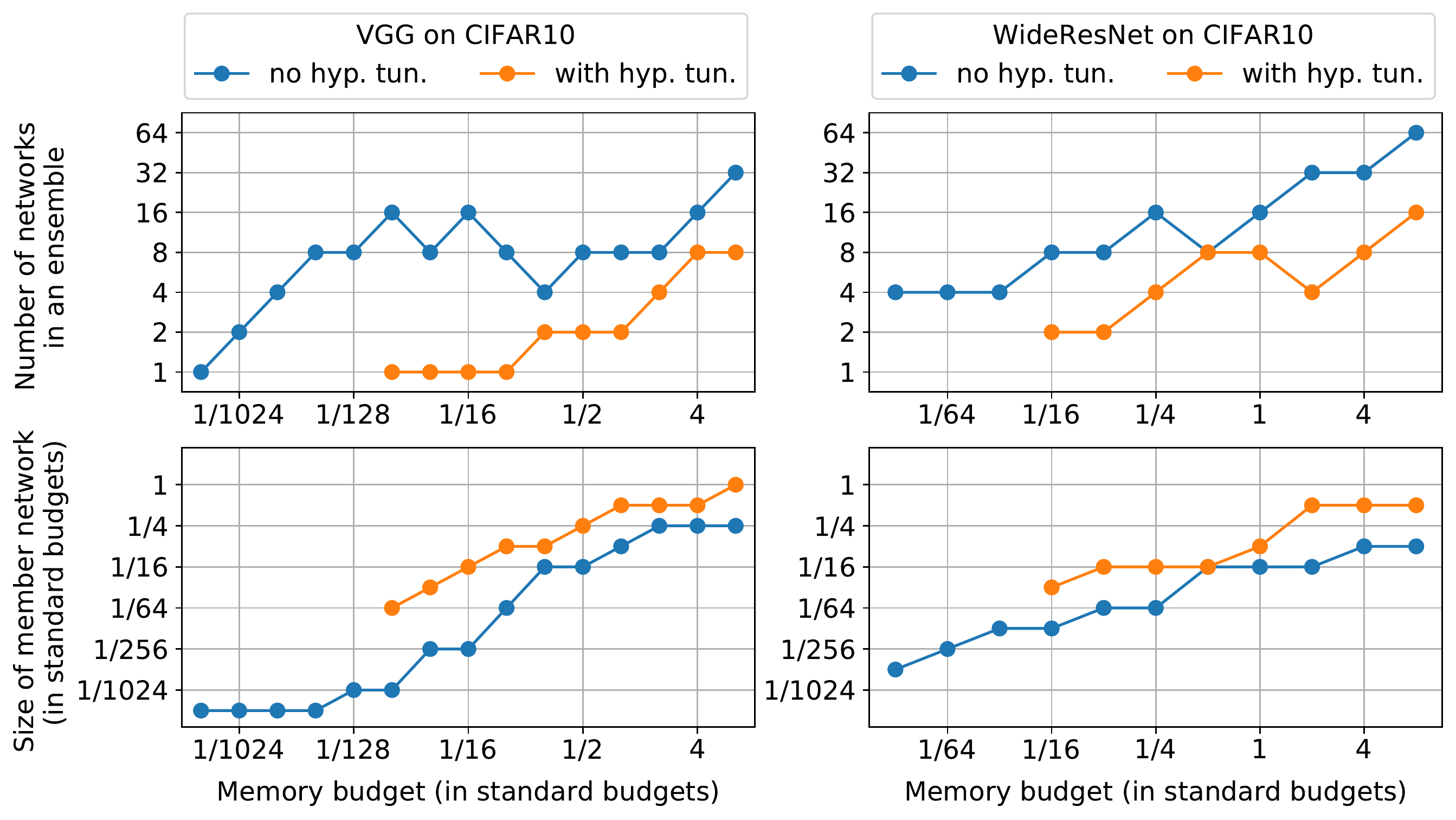} & \hspace{0.3cm} &
        \includegraphics[height=5.5cm]{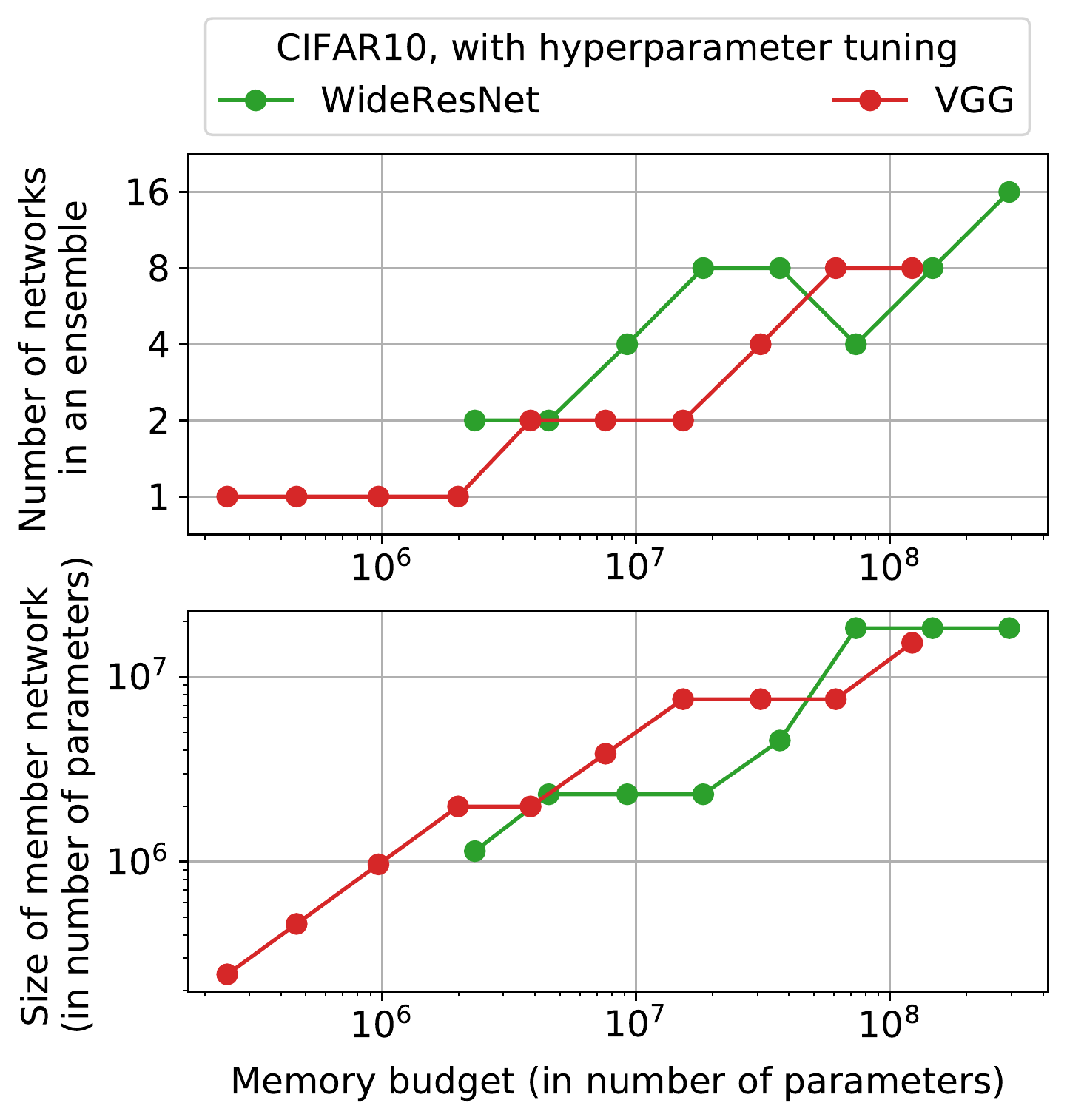}
    \end{tabular}
}
\caption{Optimal memory splits for VGG and WideResNet on CIFAR-10 for different memory budgets. For each optimal memory split the ensemble size is shown in the top plot, while the corresponding size of the member network is shown in the bottom plot. Left: results for settings with and without hyperparameter tuning. Right: comparison of memory splits for VGG and WideResNet, aligned in terms of number of parameters.}
\label{fig:opts_cifar10}
\end{center}
\vskip -0.2in
\end{figure*}

\begin{figure}[h!]
\vskip 0.05in
\begin{center}
\centerline{
    \includegraphics[height=5.5cm]{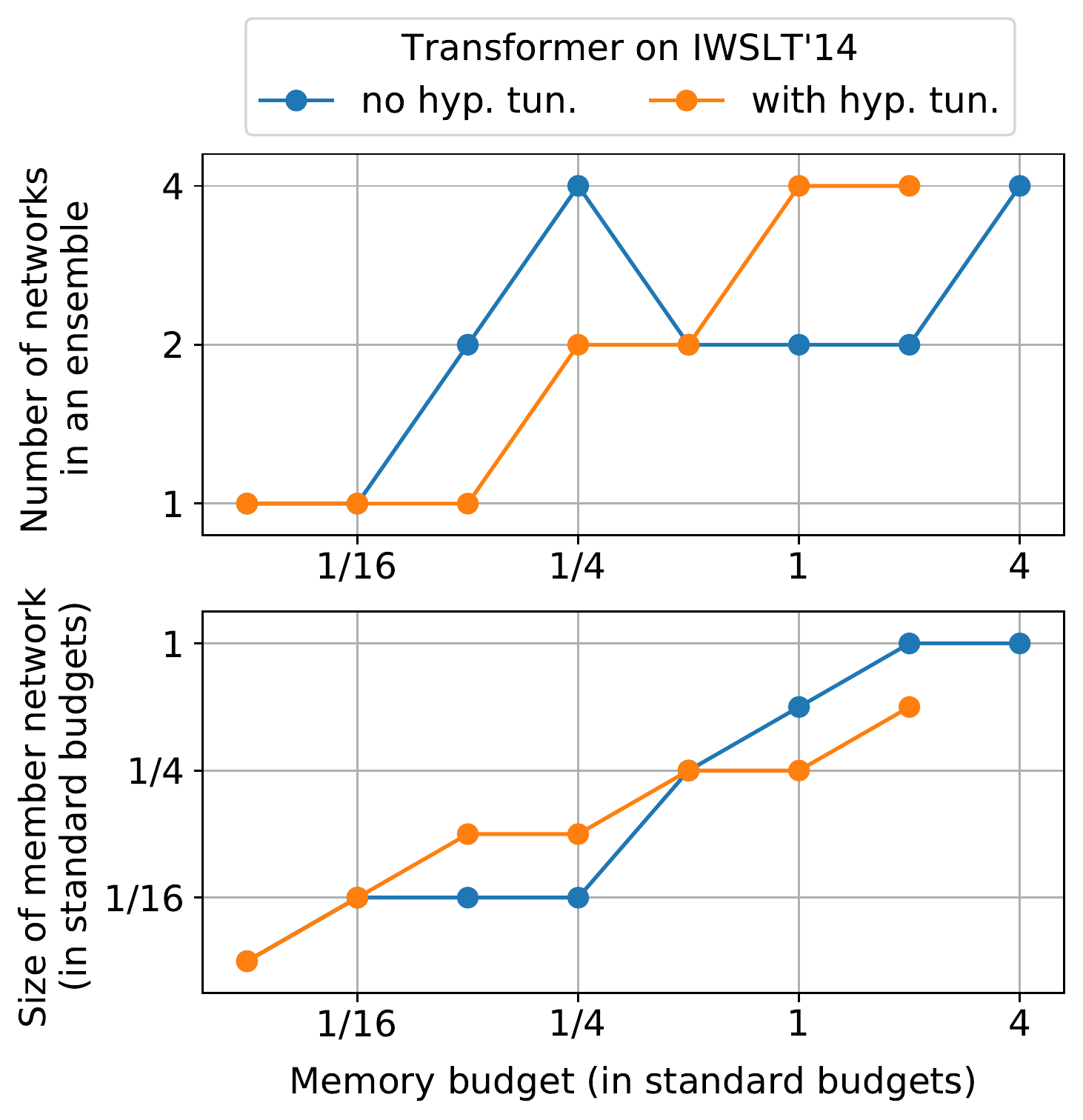} 
}
\caption{Optimal memory splits for Transformer on IWSLT'14 De-En for different memory budgets for settings with and without hyperparameter tuning. For each optimal memory split the ensemble size is shown in the top plot, while the corresponding size of the member network is shown in the bottom plot.}
\label{fig:opts_transformer}
\end{center}
\vskip -0.2in
\end{figure}

Optimal memory splits for VGG/WideResNet on CIFAR-10 and Transformer on IWSLT'14  De-En for different memory budgets are shown in figures~\ref{fig:opts_cifar10} and~\ref{fig:opts_transformer} correspondingly. 
The general pattern for these dataset--architecture pairs look very similar to results on CIFAR-100:  with the increasing memory budget, the optimal memory split grows both in terms of ensemble size and member network size.

In case of VGG and WideResNet on CIFAR-10, similar to CIFAR-100, optimal memory splits for the setting with hyperparameter tuning contain fewer networks of larger size. 
However, in case of Transformer, the comparison results are not that clear.
The reason 
is that hyperparameter tuning in case of Transformer not only makes large networks better by regularizing them, but also makes smaller networks better by employing higher learning rates.

The comparison results of optimal memory splits between VGG and WideResNet on CIFAR-10 generally looks similar to the results on CIFAR-100 but a bit more noisy (figure~\ref{fig:opts_cifar10}, right). 

\section{MSA effect for uncertainty estimation}
\label{append:uncertainty}
\begin{figure}[h!]
\begin{center}
\centerline{
    \begin{tabular}{c}
        \small{VGG, CIFAR-100, tuned hyperparameters} \\
        \includegraphics[height=4.35cm]{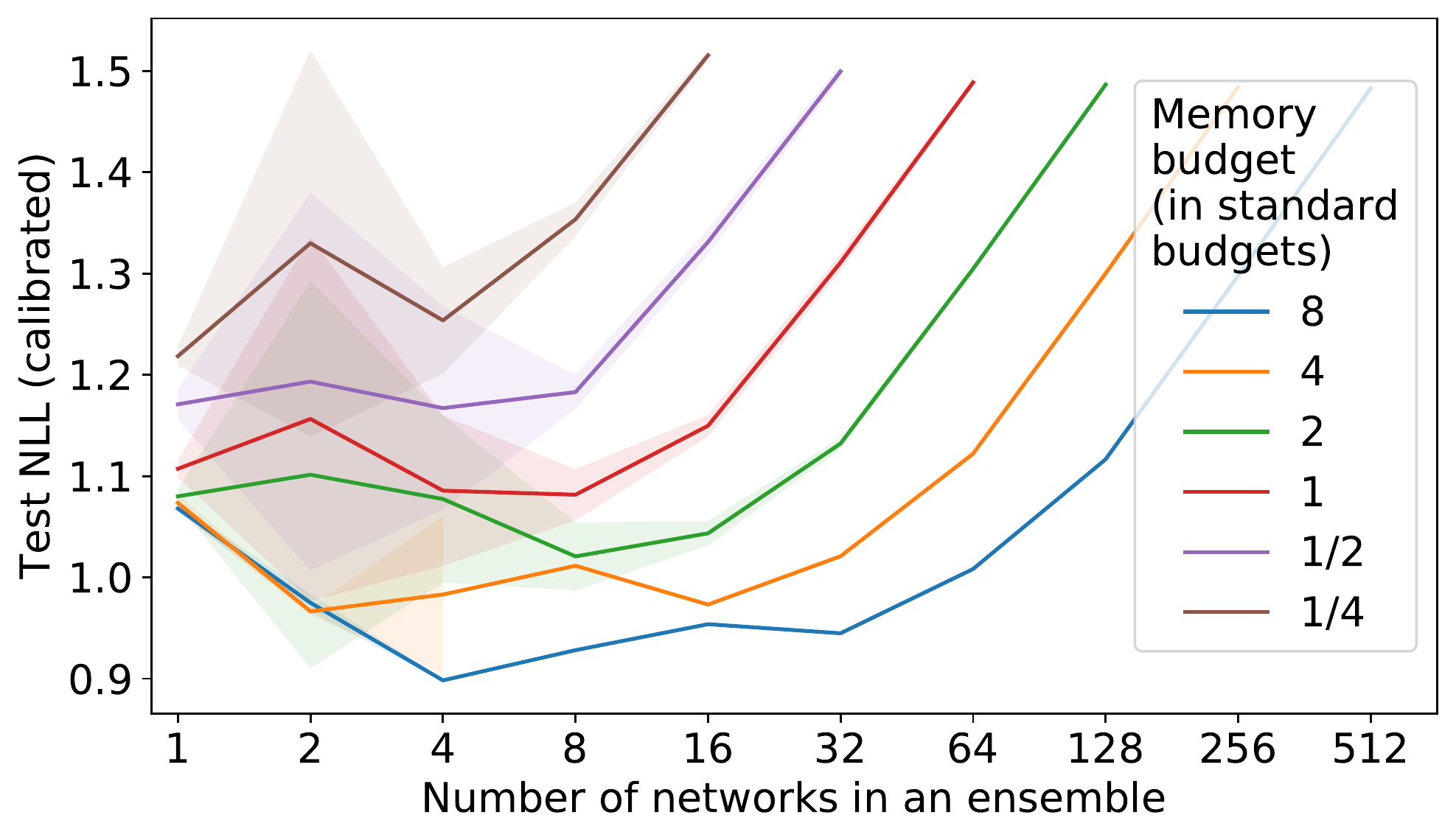}\\
        \small{VGG, CIFAR-10, tuned hyperparameters} \\
        \includegraphics[height=4.35cm]{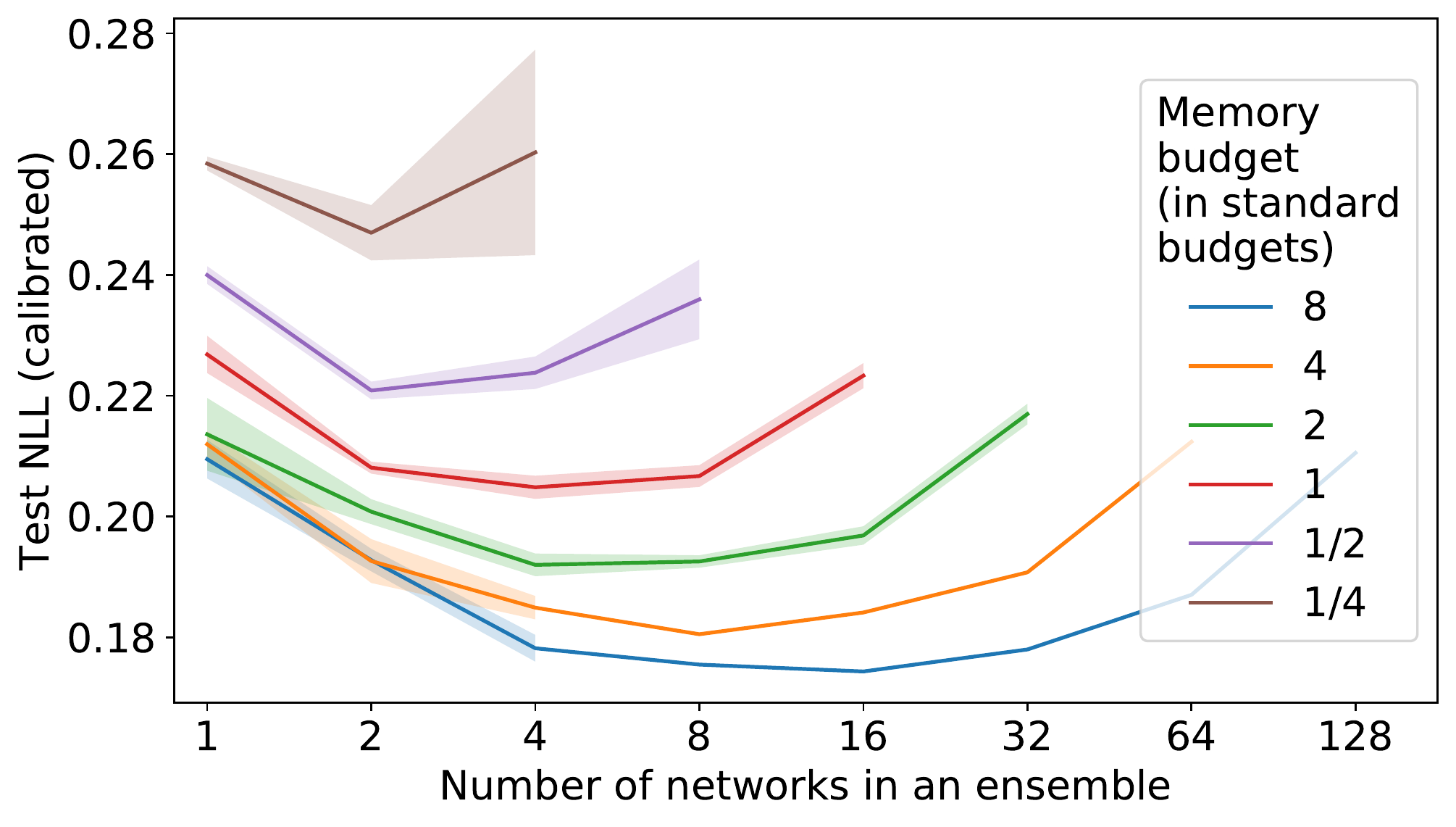} 
    \end{tabular}
}
\caption{MSA effect for calibrated test NLL, VGG on CIFAR-100 and CIFAR-10. Each line shows mean $\pm$ standard deviation of the calibrated test NLL of the ensembles with a fixed memory budget; the x-axis denotes the number of networks $N$ in the ensemble. One standard budget corresponds to the commonly used model size. For some large ensembles, the standard deviation is not provided due to the limited computational resources.}
\label{fig:uncertainty_app}
\end{center}
\vskip -0.2in
\end{figure}

Memory split plots for the calibrated test NLL for VGG with tuned hyperparameters are shown in figure~\ref{fig:uncertainty_app}. The results generally look similar to the case of WideResNet: the MSA effect holds for the calibrated test NLL in all the same cases as for the test accuracy.

\end{document}